%% file: main.tex
\documentclass{article} 
\usepackage{arxiv}

\input{math_commands.tex}

\usepackage[utf8]{inputenc} 
\usepackage[T1]{fontenc}    
\usepackage{hyperref}       
\usepackage{url}            
\usepackage{booktabs}       
\usepackage{amsfonts}       
\usepackage{bm}
\usepackage{nicefrac}       
\usepackage{microtype}      
\usepackage{lipsum}		
\usepackage{subfigure}
\usepackage{graphicx}
\usepackage[numbers]{natbib}
\usepackage{doi}
\usepackage{breqn}
\usepackage{amsmath}
\usepackage{booktabs}
\usepackage{nicefrac}
\usepackage{fancyhdr}
\newcommand{\MLP}{\texttt{MLP}}
\newcommand{\sqp}{\texttt{squareplus}}

\newcommand{\CV}{\mathcal{V}}
\newcommand{\CE}{\mathcal{E}}

\newcommand{\cW}{\mathbf{W}}

\newcommand{\cx}{\mathbf{x}}

\newcommand{\ch}{\mathbf{h}}

\newcommand{\cz}{\mathbf{z}}

\title{Discovering Symbolic Laws Directly from Trajectories with Hamiltonian Graph Neural Networks}
\author{Suresh Bishnoi \\
  School of Interdisciplinary Research,\\
  Indian Institute of Technology Delhi,\\
  New Delhi 110016, India\\
  \texttt{suresh@sire.iitd.ac.in}
  \And
  Ravinder Bhattoo \\
  Department of Civil Engineering,\\
  Indian Institute of Technology Delhi,\\
  New Delhi 110016, India\\
  \texttt{cez177518@iitd.ac.in}
  \And
  Jayadeva \\
  Department of Electrical Engineering,\\
  Indian Institute of Technology Delhi,\\
  New Delhi 110016, India\\
  \texttt{jayadeva@ee.iitd.ac.in}
  \And
  Sayan Ranu \\
  Department of Computer Science and Engineering,\\
  Yardi School of Artificial Intelligence,\\
  Indian Institute of Technology Delhi,\\
  New Delhi 110016, India\\
  \texttt{sayanranu@iitd.ac.in}
  \And
  N M Anoop Krishnan\\
  Department of Civil Engineering,\\
  Yardi School of Artificial Intelligence,\\
  Indian Institute of Technology Delhi,\\
  New Delhi 110016, India\\
  \texttt{krishnan@iitd.ac.in}
}

\begin{document}
\maketitle

\begin{abstract}
The time evolution of physical systems is described by differential equations, which depend on abstract quantities like energy and force. Traditionally, these quantities are derived as functionals based on observables such as positions and velocities. Discovering these governing symbolic laws is the key to comprehending the interactions in nature. Here, we present a Hamiltonian graph neural network (\hgnn{}), a physics-enforced \gnn{} that learns the dynamics of systems directly from their trajectory. We demonstrate the performance of \hgnn{} on $n-$springs, $n-$pendulums, gravitational systems, and binary Lennard Jones systems; \hgnn{} learns the dynamics in excellent agreement with the ground truth from small amounts of data. We also evaluate the ability of \hgnn{} to generalize to larger system sizes, and to hybrid spring-pendulum system that is a combination of two original systems (spring and pendulum) on which the models are trained independently. Finally, employing symbolic regression on the learned \hgnn{}, we infer the underlying equations relating the energy functionals, even for complex systems such as the binary Lennard-Jones liquid. Our framework facilitates the interpretable discovery of interaction laws directly from physical system trajectories. Furthermore, this approach can be extended to other systems with topology-dependent dynamics, such as cells, polydisperse gels, or deformable bodies.
\end{abstract}

Any system in the universe is always in a continuous state of motion. This motion, also known as the dynamics, is observed and noted in terms of the trajectory, which comprises the system's configuration (that is, positions and velocities) as a function of time. Any understanding humans have developed about the universe is through analyzing the dynamics of different systems. Traditionally, the dynamics governing a physical system are expressed as governing differential equations derived from fundamental laws such as energy or momentum conservation, which, when integrated, provide the system's time evolution. However, these equations require the knowledge of functionals that relate  abstract quantities such as energy, force, or stress with the configuration~\cite{rapaport2004art}. Thus, discovering these governing equations directly from the trajectory remains the key to understanding and comprehending the phenomena occurring in nature. Alternatively, several symbolic regression (SR) approaches have been used to discover free-form laws directly from observations~\cite{schmidt2009distilling,udrescu2020ai,cornelio2023combining}. However, the function space to explore in such cases is prohibitively large, and appropriate assumptions and constraints regarding the equations need to be provided to obtain a meaningful and straightforward equation~\cite{liu2021machine,liu2021machine_nonconservative,cranmer2023interpretable}. 

Learning the dynamics of physical systems directly from their trajectory is a problem of interest in wide areas such as robotics, mechanics, biological systems such as proteins, and atomistic dynamics~\citep{cranmer2020discovering,karniadakis2021physics,bapst2020unveiling,park2021accurate,battaglia2013simulation}. Recently, machine learning (ML) tools have been widely used to learn the dynamics of systems directly from the trajectory of systems~\cite{sanchez2020learning,zhong2021benchmarking,lnn,lnn1,lnn2,gruver2021deconstructing,tamayo2020predicting}. Specifically, there have been three broad approaches to this extent, namely, data-driven, physics-informed, and physics-enforced approaches. Data-driven approaches try to develop models that learn the dynamics directly from ground-truth trajectories~\cite{sanchez2020learning,bapst2020unveiling,battaglia2013simulation}. Physics-informed approaches rely on an additional term in the loss function, which is the governing differential equation: data loss and physics loss~\cite{karniadakis2021physics}. In contrast, physics-enforced approaches directly infuse the inductive biases in terms of the ordinary differential equations directly in the formulation as a hard constraint. These approaches are known as Hamiltonian (\hnn)~\citep{sanchez2019hamiltonian,greydanus2019hamiltonian,zhong2020dissipative,zhong2021benchmarking}, and Lagrangian neural networks (\textsc{Lnn})~\citep{lnn,lnn1,lnn2}, and Graph Neural ODEs~\cite{chen2018neural,gruver2021deconstructing,bishnoi2023enhancing}. Adding the inductive bias in a physics-enforced fashion instead of a soft constraint in the loss function can significantly enhance the learning efficiency while also leading to realistic trajectories in terms of conservation laws~\cite{zhong2021benchmarking,zhong2020dissipative,thangamuthuunravelling}. Additionally, combining these formulations with graph neural networks (\gnn{s})~\citep{scarselli2008graph,bhattoo2023learning,bhattoolearning,thangamuthuunravelling} can lead to superior properties such as zero-shot generalizability to unseen system sizes and hybrid systems unseen during the training, more efficient learning, and inference. However, although efficient in learning the dynamics, these approaches remain black-box in nature with poor interpretability of the learned function, which questions the robustness and correctness of the learned models~\cite{grojean2022lessons}.

Here, we present a framework combining Hamiltonian graph neural networks (\hgnn) and symbolic regression (SR), which enables the discovery of symbolic laws governing the energy functionals directly from the trajectory of systems. Specifically, we propose a \hgnn{} architecture that decouples kinetic and potential energies and, thereby, efficiently learns the Hamiltonian of a system directly from the trajectory. We evaluate our architecture on several complex systems such as $n$-pendulum, $n$-spring, $n$-particle gravitational, and binary LJ systems. Further, the modular nature of \hgnn{} enables the interpretability of the learned functions, which, when combined with SR, enables the discovery of the governing laws in a symbolic form, even for complex interactions such as binary LJ systems.

\section*{Hamiltonian mechanics}
Here, we briefly introduce the mathematical formulation of  Hamiltonian mechanics that govern the dynamics of physical systems. Consider a system of $n$ particles that are interacting with their positions at time $t$ represented by the Cartesian coordinates as $\mathbf{x}(t)=(\mathbf{x_1}(t),\mathbf{x_2}(t),...\mathbf{x_n}(t))$. The Hamiltonian $H$  of the system is defined as $H(\p,\x)=T(\dot{\x})+V(\x)$, where $T(\dot{\x})$ represents the total kinetic energy and $V(\x)$ represents the potential energy of the system. The Hamiltonian equations of motion for this system in Cartesian coordinates are given by~\citep{lavalle2006planning,goldstein2011classical,murray2017mathematical}
\begin{equation}
    {\dot{\x}} = \nabla_{\mathbf{p}_{\x}} H, \qquad {\dot{\mathbf{p}_\x}} = -\nabla_{\x} H
\end{equation}
where $\p=\nabla_{\dot{x}}H=\mathbf{M}\dot{\x}$ represents the momentum of the system in Cartesian coordinates and $\mathbf{M}$ represents the mass matrix. Assuming $Z = [\x; \p]$ and $J = [0, I; -I, 0]$, the acceleration of a particle can be obtained from the Hamiltonian equations as
\begin{equation}
    \dot{Z} = J(\nabla_Z H)
    \label{eq:ham}
\end{equation}
since $\nabla_Z H + J\dot{Z} = 0$ and ${J}^{-1} = -{J}$.
Sometimes systems may be subjected to constraints that depend on positions (holonomic) or velocities (Pfaffian). For example, in the case of a pendulum, the length between the bobs remains constant, or in multi-fingered grasping, the velocity of two fingers should be such that the combined geometry is able to hold the object. In such cases, the constrain equation is represented as $\Phi(\x)\dot{\x}=0$, where $\Phi(\x) \in \mathbb{R}^{k \times D}$ correspond to the $k$ velocity constraints in a $D$-dimensional system. For instance, in the case of a pendulum, the constraint equation for two bobs located at $(0,0)$ and $(x_1,x_2)$ may be written as $x_1\dot{x}_1 + x_2\dot{x}_2 = 0$, which is the gradient of $x_1^2 + x_2^2 =0$. Following this, the Hamiltonian equations of motion can be modified to feature the constraints explicitly as~\cite{lnn1,murray2017mathematical}
\begin{equation}
    \nabla_Z H + J\dot{Z} + (D_Z \Psi)^T\lambda = 0
\label{eq:ham_constraint}
\end{equation}
where $\Psi(Z) = (\Phi;\dot{\Phi})$, $D_Z \Psi$ is the Jacobian of $\Psi$ with respect to $Z$, and $(D_Z \Psi)^T\lambda$ represents the effect of constraints on $\dot{\x}$ and $\dot{\p}$~\cite{lnn1,murray2017mathematical}. Thus, $(D_Z\Psi)\dot{Z}=0$. Substituting for $\dot{Z}$ from Eq.~\ref{eq:ham_constraint} and solving for $\lambda$ yields~\cite{lnn2,thangamuthuunravelling,gruver2021deconstructing,lavalle2006planning}
\begin{equation}
\label{eq:lambda}
    \lambda = -[(D_Z\Psi)J(D_Z\Psi)^T]^{-1}[(D_Z\Psi)J(\nabla_Z H)]
\end{equation}

Substituting $\lambda$ in the Eq.~\ref{eq:ham_constraint} and solving for $\dot{Z}$ yields
\begin{equation}
    {\dot{Z}} = J[\nabla_Z H -(D_Z \Psi)^T [(D_Z\Psi)J(D_Z\Psi)^T]^{-1} (D_Z \Psi) J \nabla_Z H]
    \label{eq:acc_ham}
\end{equation}
Note that in the absence of constraint, Eq.~\ref{eq:acc_ham} reduces to Eq.~\ref{eq:ham}. In Hamiltonian mechanics, Eq.{\ref{eq:acc_ham}} is used to obtain the acceleration of the particles, which, when integrated, provides the updated configuration of the system. Thus, the only unknown in the previous equation is the $H$, which is represented as a function of $\p$ and $\x$. 

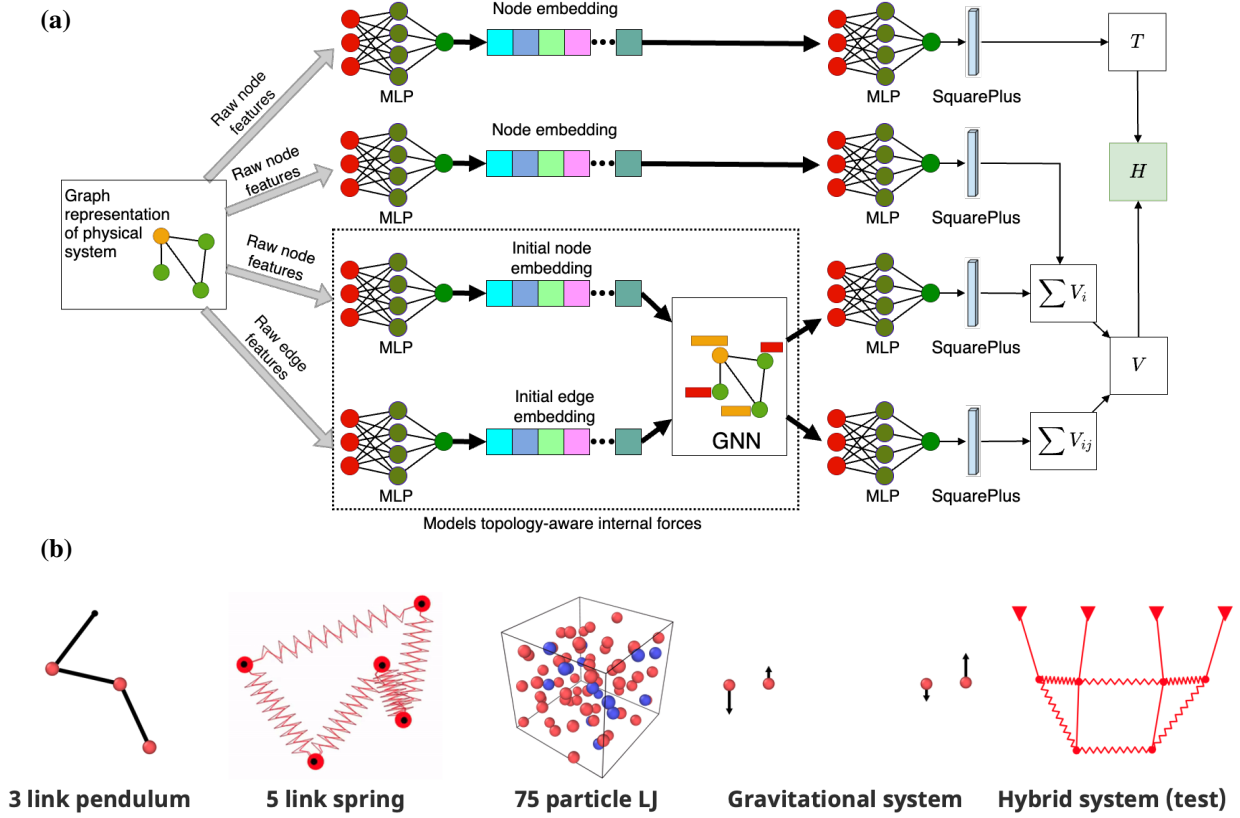
\begin{figure}
    \centering
    \begin{subfigure}
        \centering
        \includegraphics[width=0.9\columnwidth]{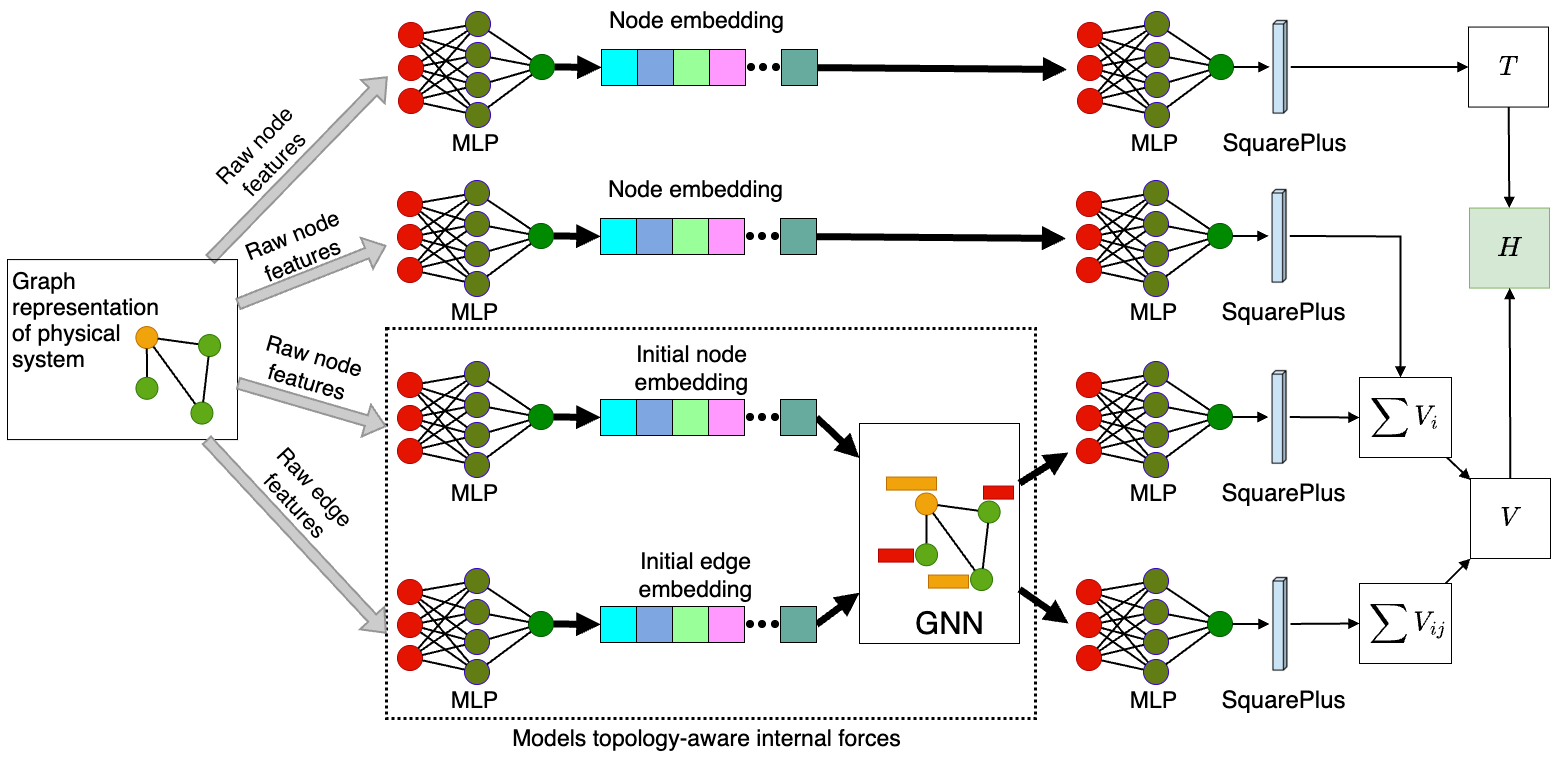}
    \end{subfigure}
    \begin{subfigure}
        \centering
        \vspace{0.20in}
        \includegraphics[width=\linewidth]{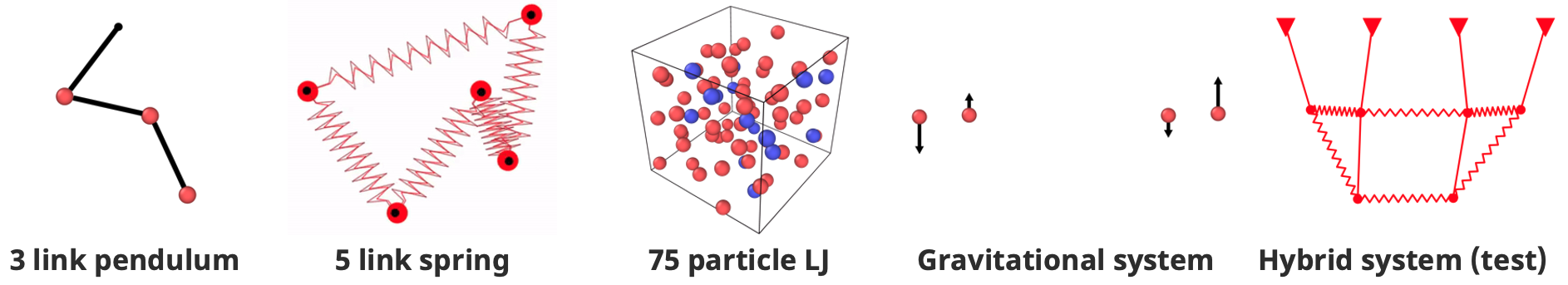}
    \end{subfigure}
    \begin{picture}(0,0)
        \put(-220,310){\textbf{(a)}}
        \put(-220,110){\textbf{(b)}}
    \end{picture}    
    \caption{\textbf{Hamiltonian graph architecture and systems studied.} (a) Hamiltonian graph neural network (\hgnn{}) architecture, (b) Visualization of the systems studied, namely, $3$-pendulum, $5$-spring, $75$-particles binary Lennard Jones system, $4$-particle gravitational system, and a hybrid spring-pendulum system. Note that the hybrid spring-pendulum system is used only to evaluate the generalizability of \hgnn{}.}
    \label{fig:arch_sys_fig}
\end{figure}

\section*{Hamiltonian graph neural network}
Now, we introduce our ML framework proposed to learn the Hamiltonian of a system directly from the trajectory, that is, only using the time evolution of the observable quantities $(\x,\p)$. To this extent, we develop the Hamiltonian graph neural network (\hgnn{}) that parametrizes the actual $H$ as a \gnn{} to obtain the learned $\hat{H}$. Henceforth, all the terms with a hat, for example, $\hat{\x}$ represent the approximate function obtained from \hgnn{}. Further, the $\hat{H}$ obtained from \hgnn{} is substituted in the Eq.(\ref{eq:acc_ham}) to obtain the acceleration and velocity of the particles. These values are integrated using a symplectic integrator to compute the updated position. 

First, we describe the architecture of \hgnn{} (see Fig.~\ref{fig:arch_sys_fig}(a)). The physical system is modeled as an undirected graph $\mathcal{G}=(\mathcal{V},\mathcal{E})$ with nodes as particles and edges as connections between them. For instance, in an $n$-ball-spring system, the balls are represented as nodes and springs as edges. The raw node features are $t$ (type of particle) as one-hot encoding, $\x$, and $\dot{\x}$, and the raw edge feature is the distance, $d=||\x_j-\x_i||$, between two particles $i$ and $j$. A notable difference in the \hgnn{} architecture from previous graph architectures is the presence of global and local features---local features participate in message passing and contribute to quantities that depend on topology. In contrast, global features do not take part in message passing. Here, we employ the position $\x$, velocity $\dot{\x}$ as global features for a node, while $d$ and $t$ are used as local features.

For the \gnn{}, we employ an $L$-layer message passing \gnn{}, which takes an embedding of the node and edge features created by multi-layer perceptrons (MLPs) as input. Detailed hyper-parameters are provided in the Supplementary Material. The local features participate in message passing to create an updated node and edge embeddings. The final representations of the nodes and edges, $\cz_i$ and $\cz_{ij}$, respectively, are passed through MLPs to obtain the Hamiltonian of the system. The Hamiltonian of the system is predicted as the sum of kinetic energy $T$ and potential energy $V$ in the \hgnn{}. Specifically, the potential energy is predicted as $V_i=\sum_i \mlp_v(\cz_i)+ \sum_{ij} \mlp_e(\cz_{ij}))$, where $\mlp_v$ and $\mlp_e$ represent the contribution from the node (particles themselves) and edges (interactions) toward the potential energy of the system, respectively. Kinetic energy is predicted as $T= \sum_i \mlp_T\left(\ch^0_i\right)$, where $\ch^0_i$ is the embedding of particle $i$.

To train the \hgnn{}, we use only the time evolution of positions and momenta. This approach does not assume any knowledge of the functional form or knowledge of the Hamiltonian. The training approach, purely based on direct observables, can be used for any system (for example, trajectories from experiments) where the true Hamiltonian is unavailable. Thus, the loss function of \hgnn{} is computed by using the predicted and actual positions at the timestep $t+1$ in a trajectory based on positions and velocities at $t$, which is then back-propagated to train the MLPs. Specifically, we use \textit{mean squared error (MSE)} on the true and predicted $Z$, which is the concatenation of positions and velocities.
\begin{equation}
    \label{eq:lossfunction}
\mathcal{L}= \frac{1}{n}\left(\sum_{i=1}^n \left({Z}_i^{t+1}-\hat{{Z}}_i^{t+1}\right)^2\right)
\end{equation}

\section*{Case studies}
\textbf{Systems studied.} Now, we evaluate the ability of \hgnn{} to learn the dynamics directly from the trajectory. To evaluate \hgnn{}, we selected four different types of systems, \textit{viz}, $5$-pendulums with explicit internal constraints and subjected to an external gravitational field, $5$-springs with harmonic inter-particle harmonic interactions, $75$-particle binary LJ system with two types of a particle interacting based on the Kob-Andersen LJ potential~\cite{kob1994scaling}, and $4$-particle gravitational system with purely repulsive gravitational potential. Finally, in order to test the generalizability of \hgnn{} to completely unseen system which is combination two systems on which it is trained, a hybrid system containing spring and pendulum is also considered. In this system, while the dynamics of pendulum is governed by the external gravitational field, the dynamics of the spring system depends on the internal forces generated in the system due to the expansion and compression of the spring. Thus, the systems selected here covers a broad range of cases, that is, dynamics (i) with internal constraints (pendulum), (ii) under the influence of an external field (gravitational), (iii) harmonic interactions (springs), (iv) complex breakable interactions (LJ potential), and (v) hybrid system with and without internal constraints.  

The training of \hgnn{} is carried out for each system separately. A training dataset of $100$ trajectories, each having $100$ steps, were used for each system. For spring and pendulum, a 5-particle system is considered with random initial conditions. In the pendulum system, the initial conditions are considered in such a fashion that the constraints are respected. In the spring system, each ball is connected only to two other balls forming a loop structure. For gravitational system, a 4-particle system is considered where two particles are rotating in the clockwise direction, and two remaining particles are rotating in the anti-clockwise direction about their center of mass. For LJ system, a binary Kob-Andersen system with 75 particles are considered. The initial structure is generated by randomly placing the particles in a box with periodic boundary conditions. Further, the systems are simulated in a microcanonical ensemble (NVE) with temperatures corresponding to the liquid state to obtain equilibrium structures. Only once the system is equilibrated, the training data is collected for this system. \hgnn{} models were trained on this dataset with a $75:25$ split for training and validation. Further, to test the long-term stability and energy and momentum conservation error, the trained model was evaluated on a forward simulation for 10$^5$ timesteps on 100 random initial configurations. See Methods for detailed equations for the interactions, datasets, and training parameters. 

\begin{figure}
\centering
  \includegraphics[width=\linewidth]{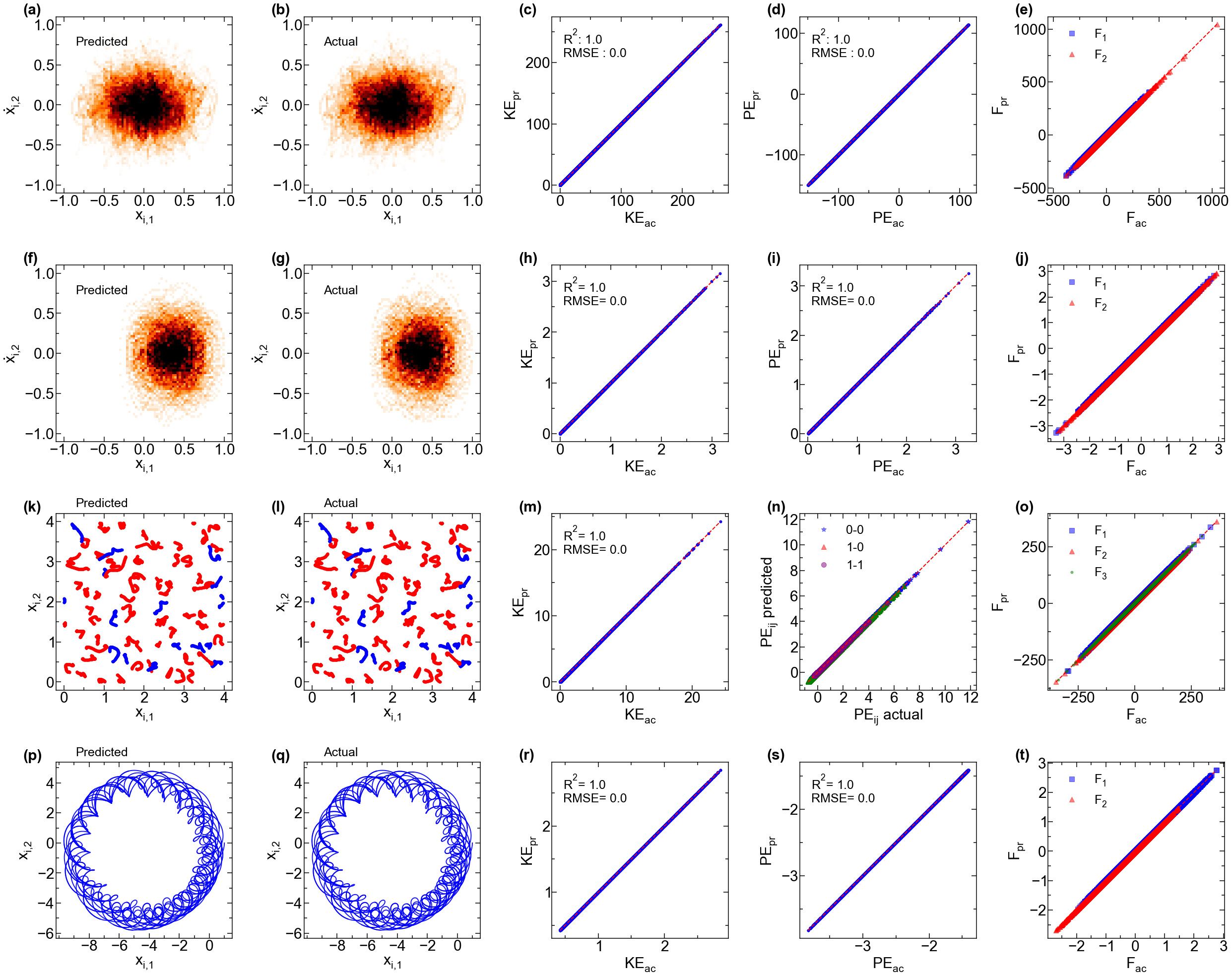}
  \caption{\textbf{Evaluation of \hgnn{} on the pendulum, spring, binary LJ, and gravitational systems.} (a) Predicted and (b) actual phase space (that is, $x_1$-position vs. $x_2$-velocity), predicted with respect to actual (c) kinetic energy, (d) potential energy, and (e) forces in 1 (blue square), and 2 (red triangle) directions of the $5$-pendulum system. (f) Predicted and (g) actual phase space (that is, $1$-position, $x_1$ vs $2$-velocity, $\dot{x}_2$), predicted with respect to actual (h) kinetic energy, (i) potential energy, and (j) forces  in 1 (blue square) and 2 (red triangle) directions of the $5$-spring system. (k) Predicted and (l) actual positions (that is, $x_1$ and $x_2$ positions), predicted with respect to actual (m) kinetic energy, (n) pair-wise potential energy, $V_{\textrm{ij}}$ for the (0-0), (0-1), and (1-1) interactions, and (o) forces  in 1 (blue square), 2 (red triangle), and 3 (green circle) directions of the $75$-particle LJ system. (p) Predicted and (q) actual positions (that is, $x_1-$ and $x_2-$positions), predicted with respect to actual (r) kinetic energy, (s) potential energy, and (t) forces in 1 (blue square), and 2 (red triangle) directions of the gravitational system.}
  \label{fig:fig2_pend_system}
\end{figure}

\textbf{Learning the dynamics.} Now, we evaluate the performance of the trained \hgnn{} models. To evaluate the long-term stability of the dynamics learned by \hgnn{}, we analyze the trajectory predicted by \hgnn{} for 100 random initial configurations. Specifically, we compare the predicted and actual phase space, trajectory, kinetic energy, potential energy, and forces on all the particles of the system during the trajectory. Note that the systems studied in this case are chaotic; hence, the exact trajectory followed by \hgnn{} will diverge with time. However, the phase space and the errors in energy and forces can be effective metrics to analyze whether the trajectory generated by \hgnn{} is statistically equivalent to that of the original system, that is, sampling the same regions of the energy landscape. Further, in contrast to purely data-driven~\cite{cranmer2020discovering} or physics-informed methods, the physics-enforced architecture of \hgnn{} strictly follows all the characteristics of the Hamiltonian equations of motion, such as the conservation laws of energy and momentum (see Supplementary Materials). This is due to the fact that the graph architecture only predicts the Hamiltonian of the system, which is then substituted in the Hamiltonian equations of motion to obtain the updated configuration. Due to this feature, the trajectory predicted by the \hgnn{} is more realistic and meaningful in terms of the system's underlying physics.

Fig.~\ref{fig:fig2_pend_system} shows the performance of \hgnn{} for the pendulum (Figs.~\ref{fig:fig2_pend_system}(a)-(e), first row), spring (Figs.~\ref{fig:fig2_pend_system}(f)-(j), second row), binary LJ (Figs.~\ref{fig:fig2_pend_system}(k)-(o), third row), and gravitational systems (Figs.~\ref{fig:fig2_pend_system}(p)-(t), fourth row). For pendulum and spring systems, we observe that the phase space represented by the positions in 1-direction ($x_1$) and velocities in the orthogonal direction ($\dot{x}_2$) predicted by \hgnn{} (Figs.~\ref{fig:fig2_pend_system}(a) and (f)) exhibit an excellent match with the ground truth trajectory. It is interesting to note that \hgnn{} trained only on a trajectory of a single step ($t$ to $t+1$) is able to learn the dynamics accurately and simulate a long-term stable trajectory of $10^5$ timesteps that exactly matches the simulated trajectory. Similarly, for the binary LJ and gravitational systems, we observe that the predicted (Figs.~\ref{fig:fig2_pend_system}(k) and (p)) and actual (Figs.~\ref{fig:fig2_pend_system}(j) and (q)) positions in the trajectory of random unseen initial configurations explored by the systems exhibit an excellent match. Further, we observe that the predicted kinetic (Figs.~\ref{fig:fig2_pend_system}(c), (h), (m), (r)) and potential (Figs.~\ref{fig:fig2_pend_system}(d), (i), (n), and (s)) energies and forces (Figs.~\ref{fig:fig2_pend_system}(e), (j), (o), and (t)) exhibit an excellent match with the ground truth values with a mean squared error almost close to zero. Additional evaluation of the \hgnn{} architecture is performed by comparing it with two baselines, namely, \hnn{} (which is a physics-enforced MLP) and \hgn{}, which does not decouple potential and kinetic energies (see Supplementary Materials) and on additional metrics such as energy and momentum error. We observe that \hgnn{} significantly outperforms \hgn{} and \hnn{} in terms of rollout and energy error (see Supplementary Materials). These results confirm that the \hgnn{} architecture can learn the systems' dynamics directly from the trajectory and hence can be used for systems where the Hamiltonian is unknown or inaccessible (such as experimental or coarse-grained systems). 

\begin{figure}
\centering
  \includegraphics[width=\linewidth]{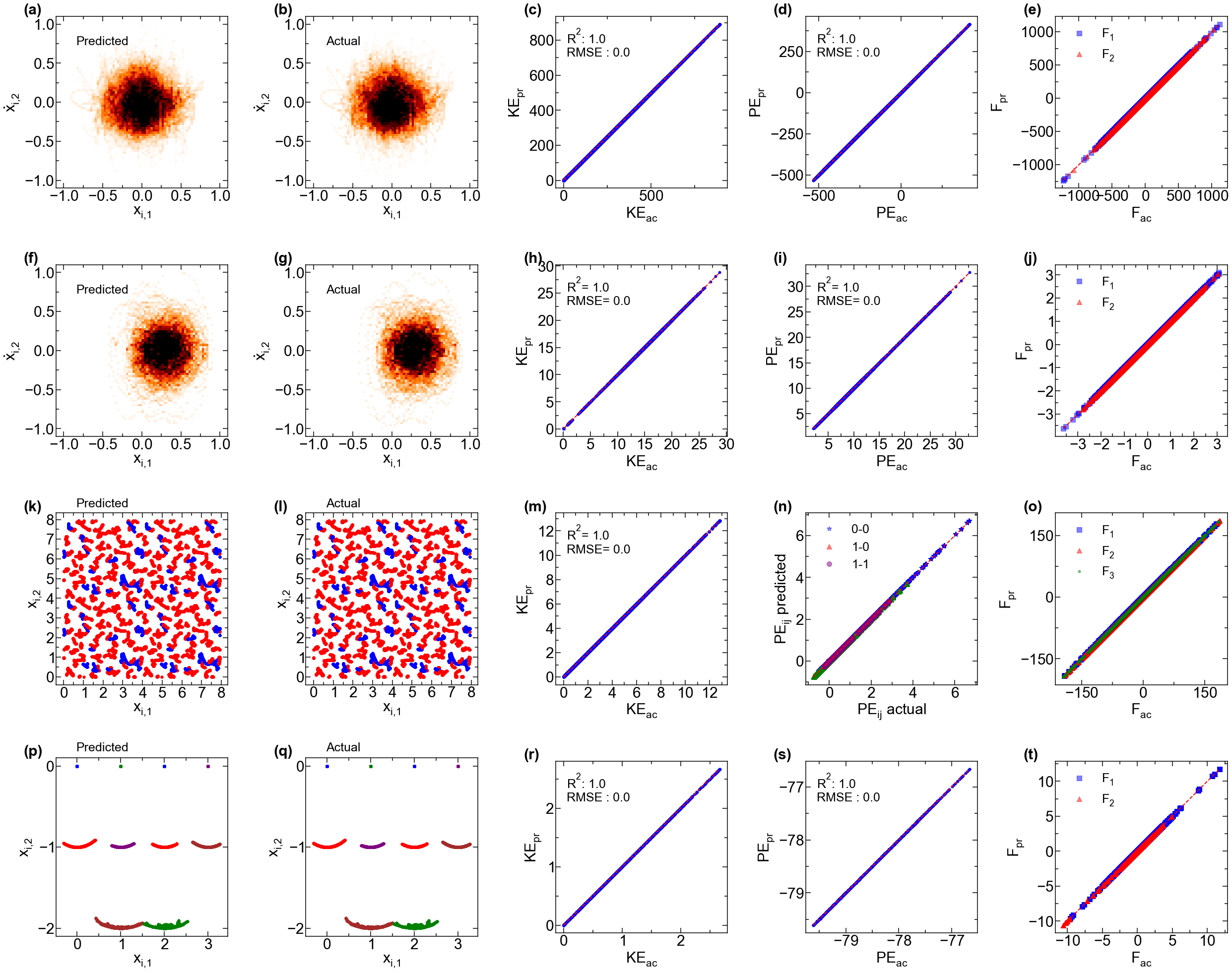}
  \caption{\textbf{Generalizability to unseen systems.} (a) Predicted and (b) actual phase space (that is, $1-$position vs. $2-$velocity) and predicted with respect to actual (c) kinetic energy, (d) potential energy, and (e) forces of the $10-$pendulum system, using \hgnn{} trained on $5-$pendulum system. (f) Predicted and (g) actual phase space (that is, $1-$position, $x_1$ vs. $2-$velocity, $\dot{x}_2$) and predicted with respect to actual (h) kinetic energy, (i) potential energy, and (j) forces of the $50-$spring system using \hgnn{} trained on $5-$spring system. (k) Predicted and (l) actual positions (that is, $1-$ and $2-$positions; blue and red represent type 0 and type 1 particles), and predicted with respect to actual (m) kinetic energy, (n) pair-wise potential energy, V$_{\textrm{ij}}$ and (o) forces, of the $600-$particle binary LJ system, using \hgnn{} trained on $75$ particle binary LJ system. (p) Predicted and (q) actual positions (that is, $1-$ and $2-$positions), and predicted with respect to actual (r) kinetic energy, (s) potential energy, and (t) forces of the 10-particle hybrid system, using \hgnn{} trained on $5$-spring and $5$-pendulum system.}
  \label{fig:fig3_spring_sys}
  \vspace{-0.20in}
\end{figure}

\textbf{Zero-shot generalizability.} Now, we evaluate the generalizability of the \hgnn{} to unseen systems, for instance,  systems larger than those on which \hgnn{} is trained or a completely new system that is a combination of two systems on which it is independently trained. While traditional neural networks based on approaches are restricted to the system sizes on which it is trained, \hgnn{} is inductive to larger (and smaller) systems than those on which they are trained. This is due to the modular nature of the \hgnn{}, thanks to the graph-based approach, where the learning occurs at the node and edge level. Fig.~\ref{fig:fig3_spring_sys} shows the generalizability of \hgnn{} to larger system sizes than those on which it is trained. Specifically, we evaluate \hgnn{} on $10-$pendulum (Fig.~\ref{fig:fig3_spring_sys}(a)-(e)), $50-$spring (Fig.~\ref{fig:fig3_spring_sys}(f)-(j)), and $600-$particle binary LJ systems (Fig.~\ref{fig:fig3_spring_sys}(k)-(o)). We observe that \hgnn{} is able to generalize to larger system sizes accurately without any additional training or fine-tuning, exhibiting excellent match with the ground truth trajectory in terms of positions, energies, and forces. Additional results on $50$-pendulum systems and $500$-spring systems are included in the Supplementary Material.

We also evaluate the ability of \hgnn{} to simulate a hybrid spring-pendulum system (see Fig.~\ref{fig:arch_sys_fig}(b) Hybrid system). To this extent, we model the Hamiltonian of the hybrid as the superposition of the Hamiltonian of spring and pendulum systems. Further, we model two graphs based on the spring and pendulum elements and use the \hgnn{} trained on the spring and pendulum systems to obtain the Hamiltonian of the system. Fig.~\ref{fig:fig3_spring_sys}(p)-(t) shows the performance of \hgnn{} on the hybrid system. \hgnn{} provides the dynamics in excellent agreement with the ground truth for the unseen hybrid system as well in terms of positions, energies, and forces. Additional results on the force predicted on each particle by \hgnn{} in comparison to the ground truth for a trajectory of 100 steps is shown in Supplementary Material. These results confirm that \hgnn{} is able to learn the dynamics of systems directly from their trajectory and simulate the long-term dynamics for new initial conditions and system sizes. This is a highly desirable feature as \hgnn{} can be used to learn the Hamiltonian from sparse experimental data of physical systems or \textit{ab-initio} simulations of atomic systems. This learned model can then be used to simulate larger system sizes to investigate phenomena with higher length scales. 

\section*{Interpretability and discovering symbolic laws}
\begin{figure}
\centering
  \includegraphics[width=\linewidth]{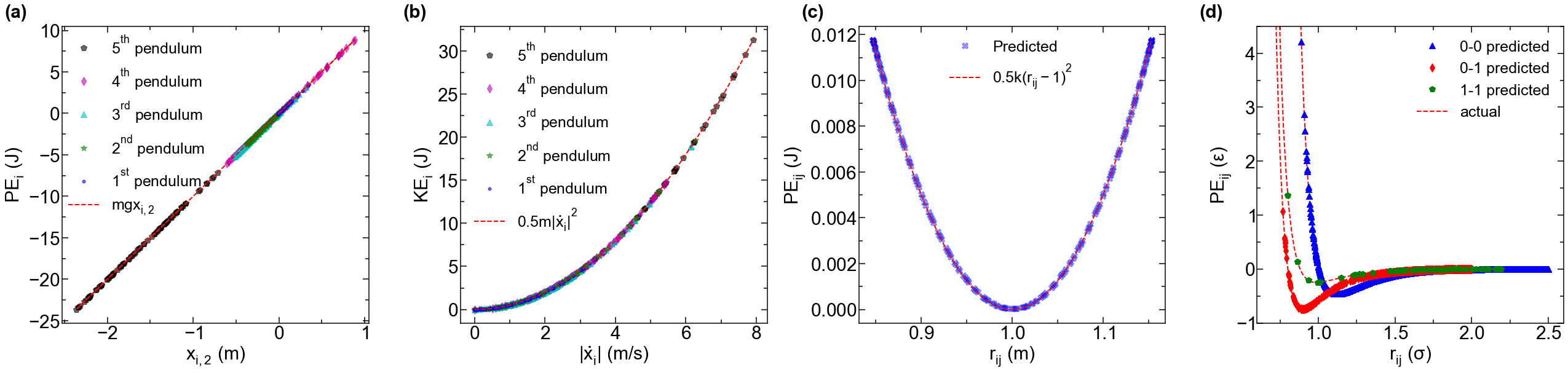}
  \caption{\textbf{Interpreting the learned functions in \hgnn{}.} (a) Potential energy of pendulum system with the $2$-position of the bobs. (b) Kinetic energy of the particles with respect to the velocity for the pendulum bobs. (c) Potential energy with respect to the pair-wise particle distance for the spring system. (d) The pair-wise potential energy of the binary LJ system for 0-0, 0-1, and 1-1 type of particles. The results from \hgnn{} are shown with the markers, while the original function is shown as dotted lines.}
  \label{fig:fig4interp}
\end{figure}

Neural networks, while exhibiting excellent capability to learn functions, are notorious for their black-box nature allowing poor or no interpretability to the learned function. In contrast, we demonstrate the interpretability of the learned \hgnn{}. Thanks to the modular nature of \hgnn{}, we analyze the functions learned by the individual MLPs that represent the node and edge level potential energies ($\mlp_v$ and $\mlp_e$, respectively) and kinetic energy ($\mlp_T$) of the particles as a function of the learned embeddings. Fig.~\ref{fig:fig4interp}(a)-(f) show the learned functions with respect to the input features such as positions, velocities, or inter-particle distances. We observe that learned functions by \hgnn{} for the potential energies for (i) pendulum bob ($mgx_2$; Fig.~\ref{fig:fig4interp}(a)), (ii) spring ($0.5k(r_{ij}-1)^2$; Fig.~\ref{fig:fig4interp}(c)), and (iii) binary LJ systems (0-0, 0-1, 1-1; Figs.~\ref{fig:fig4interp}(d)-(f), respectively) and kinetic energy of particles ($0.5m|\dot{\x}_i|^2$; Fig.~\ref{fig:fig4interp}(b)) exhibits a close match with the known governing equations. This shows the interpretability of the \hgnn{} and the additional ability to provide insights into the nature of interactions between the particles directly from their trajectory. Thus, \hgnn{} can be used to discover interaction laws directly from their trajectory, even when they are not accessible or available.

\begin{table}[h]
\begin{center}
\begin{tabular}{@{}cccccc@{}}
\toprule
Functions & Original Eq. & Discovered Eq. & Loss & Score \\
\midrule
Kinetic energy & $T_i = 0.5m|\dot{\x}_i|^2$ & $ T_i = 0.500m|\dot{\x}_i|^{2}$ & $7.96 \times 10^{-10}$ & $22.7$ \\
Harmonic spring & $V_{ij}= 0.5(r_{ij}-1)^2$ & $V_{ij} = 0.499 \left(r_{ij} - 1.00\right)^{2}$ & $1.13 \times 10^{-9}$ & $3.15$ \\
Binary LJ (0-0) & $V_{ij}=\left(\frac{2.0}{r_{ij}^{12}}-\frac{2.0}{r_{ij}^{6}}\right)$ &  $V_{ij}= \left(\frac{1.90}{r_{ij}^{12}}-\frac{1.95}{r_{ij}^{6}}\right)$ & $0.00159$ & $2.62$ \\
Binary LJ (0-1) & $V_{ij}=\left(\frac{0.275}{r_{ij}^{12}}-\frac{0.786}{r_{ij}^{6}}\right)$ & $V_{ij} = \left(\frac{2.33}{r_{ij}^{9}} - \frac{2.91}{r_{ij}^8}\right)$ & $3.47 \times 10^{-5}$ & $5.98$ \\
Binary LJ (1-1) & $V_{ij}=\left(\frac{0.216}{r_{ij}^{12}}-\frac{0.464}{r_{ij}^{6}}\right)$ & $V_{ij} = \left(\frac{0.215}{r_{ij}^{12}}-\frac{0.464}{r_{ij}^{6}}\right)$ & $1.16 \times 10^{-5}$ & $5.41$ \\
\bottomrule
\end{tabular}
\end{center}
\caption{\label{tab:sym_regression}\textbf{Discovering governing laws with symbolic regression.} Original equation and the best equation discovered by symbolic regression based on the score for different functions. The loss represents the mean squared error between the data points from \hgnn{} and the predicted equations.}
\end{table}

While the interpretability of \hgnn{} can provide insights into the nature of energy functionals, abstracting it further as a symbolic expression can enable discovering the underlying interaction laws and energy functions. Such functionals can then be used for simulating the system or understanding the dynamics independent of the \hgnn{}. Thus, beyond learning the dynamics of systems, \hgnn{} can be used to discover underlying energy functionals and interaction laws. To this extent, we apply SR~\cite{schmidt2009distilling,udrescu2020ai,cornelio2023combining} on the learned functions by \hgnn{}. Specifically, we focus on the kinetic energy function, the harmonic function of the spring, gravitational potential, and the binary LJ systems. Specifically, we employ simple operations such as addition, multiplication, and polynomials to identify the governing equations that minimize the error between the values predicted by the discovered equation and those predicted by the \hgnn{}. The optimal equation is identified based on a score that balances complexity and loss of the equation (see Methods for details).

Table~\ref{tab:sym_regression} shows the original equation and the equation discovered based on SR of the learned \hgnn{} functionals. Note for each system, the equation that exhibits the maximum score is chosen as the final equation (see Methods for details). All the equations discovered by SR with their loss, complexity, polynomials used, and other hyper-parameters are included in the Supplementary material. We observe that the recovered equations exhibit a close match for kinetic energy, harmonic spring, gravitational potential, and binary LJ. In the case of the binary LJ system, we observe that the equations reproduced for (0-0) and (1-1) interactions are very close to the original equation, while for (0-1) interaction, the equation is slightly different, although it exhibits low loss. Interestingly, we observe that for LJ (0-1) interaction, one of the equations provided by SR given by $V_{ij} = \left(\frac{0.203}{r_{ij}^{12}} - \frac{0.773}{r_{ij}^{6}}\right) $ is closer to the original equation in its functional form. However, this predicted equation has a score of $2.22$ with a loss of $0.000109$. Thus, both the loss and the score of the equation are higher and lower, respectively, than the best equation obtained in Table~\ref{tab:sym_regression}. This also suggests that for more complex interactions, an increased number of data points, especially along the inflection points, might be required to improve the probability of discovering the original equation. 
\section*{Outlook}
Altogether, in this work, we present a framework \hgnn{} that allows the discovery of energy functionals directly from the trajectory of physical systems. The \hgnn{} could be extended to address several challenging problems where the dynamics depends on the topology such as the dynamics of polydisperse gels~\cite{prost2015active}, granular materials~\cite{kou2017granular}, biological systems such as cells~\cite{trepat2018mesoscale}, or even rigid body dynamics. A topology to graph mapping can be developed in such cases which can then be used to learn the dynamics and further abstracted it out in terms of the governing interaction laws. At this juncture, it is worth mentioning some outstanding questions the present work raises. Although \hgnn{} presents a promising approach, it is applied to only particle-based systems with at most two-body interactions. Extending \hgnn{} to more complex systems, such as complex atomic structures with multi-body interactions or to deformable bodies in continuum mechanics could be addressed as future challenges. Further, the graph architecture presented in \hgnn{} could be enhanced by adding additional inductive biases such as equivariance~\cite{batzner20223}. Finally, extending the framework to non-Hamiltonian systems such as colloidal systems~\cite{li2020anatomy} exhibiting Brownian or Langevin dynamics could be pursued to widen the scope of the \hgnn{} framework to capture realistic systems.


\section*{Methods}
\subsection*{Experimental systems}
\label{app:experiments}
To simulate the ground truth, physics-based equations derived using Hamiltonian mechanics are employed. The equations for $n$-pendulum and spring systems are given in detail below.
\subsection*{$n$-Pendulum}
For an $n$-pendulum system, $n$-point masses, representing the bobs, are connected by rigid (non-deformable) bars. These bars, thus, impose a distance constraint between two point masses as 
\begin{equation}
||\x_{i}-\x_{i-1}||^2 = l_i^2
\end{equation}
where, $l_i$ represents the length of the bar connecting the $(i-1)^{th}$ and $i^{th}$ mass. This constraint can be differentiated to write in the form of a \textit{Pfaffian} constraint as
\begin{equation}
    (\x_i-\x_{i-1})(\dot{\x}_i-\dot{\x}_{i-1})=0
\end{equation}
Note that such constraint can be obtained for each of the $n$ masses considered to obtain the constraint matrix.

The Hamiltonian of this system can be written as
\begin{equation}
    H=\sum_{i=1}^n \sum_{j=1}^2\left(1/2m_i\dot{x_{i,j}}^2-m_igx_{i,2}\right)
\end{equation}
where $j=1,2$ represents the dimensions of the system, $m_i$ represents the mass of the $i^{th}$ particle, $g$ represents the acceleration due to gravity in the $2-$direction and $x_{i,2}$ represents the position of the $i^{th}$ particle in the $2-$ direction. Here, we use $l_i = 1.0$ m, $m_i = 1.0$ kg, and $g = 10.0$  $m/s^2$.

\subsection*{$n$-spring system}
Here, $n$-point masses are connected by elastic springs that deform linearly (elastically) with extension or compression. Note that similar to the pendulum setup, each mass $m_i$ is connected to two masses $m_{i-1}$ and $m_{i+1}$ through springs so that all the masses form a closed connection. The Hamiltonian of this system is given by
\begin{equation}
    H=\sum_{i=1}^n \sum_{j=1}^2\left(1/2m_i\dot{x_{i,j}}^2\right) - \sum_{i=1}^n 1/2k(||\x_{i-1}-\x_{i}||-r_0)^2
\end{equation}
where $r_0$ and $k$ represent the undeformed length and the stiffness, respectively, of the spring, and $j=1,2$ represents the dimensions of the system. Here, we use $r_0 = 1.0$ m, $m_i = 1.0$ kg and $k = 1.0$ N/m.
\subsection*{$n$-body gravitational system}
Here, $n$ point masses are in a gravitational field generated by the point masses themselves. The Hamiltonian of this system is given by
\begin{equation}
    H=\sum_{i=1}^n \sum_{j=1}^2\left(1/2m_i\dot{x_{i,j}}^2\right) + \sum_{i=1}^n\sum_{k=1,j\neq i}^n Gm_im_j/(||\x_{i}-\x_{j}||)
\end{equation}
where $G$ represents the gravitational constant, and $j=1,2$ represents the dimension of the system. Here, we use $G = 1.0$ Nm$^2$kg$^{-2}$, $m_i = 1.0$ kg and $m_j = 1.0$ kg $\forall \ i,j$.

\subsection*{Binary Lennard Jones system}
Here, we consider a binary LJ system known as the Kob-Andersen mixture~\cite{kob1994scaling} composed of 80\% particles of type 0 and 20\% particles of type 1. The particles in this system interact based on a 12-6 LJ potential with the pair-wise potential energy $V_{ij} $given by
\begin{align}
    V_{ij} = \epsilon \left[\left(\frac{\sigma}{r_{ij}}\right)^{12}-\left(\frac{\sigma}{r_{ij}}\right)^6\right]
\end{align}
where $r_{ij} = ||\x_{i}-\x_{j}||$ and $\sigma$ and $\epsilon$ are the LJ parameters, which takes the values as $\epsilon_{0-0}=1.0, \epsilon_{0-1}=1.5, \epsilon_{1-1}=0.5$ and $\sigma_{0-0}=1.00, \sigma_{0-1}=0.80, \sigma_{1-1}=0.88$, and $r_{ij}$ represents the distance between particles $i$ and $j$. The pair-wise interaction energy between all the particles is summed to obtain the total energy of the system. For the LJ system, all the simulations are conducted at a temperature of 1.2 in the microcanonical (NVE) ensemble, ensuring the system is in a liquid state. The system is initialized with atoms placed in random positions avoiding overlap in a cubic box with periodic boundary conditions with box size $3.968$ and cutoff for atom type $0-0 = 2.5$, $0-1 = 2.0$ and $1-1 = 2.2$. Further, the system is equilibrated in the NVE ensemble until the memory of the initial configuration is lost. The equations of motion are integrated with the velocity Verlet algorithm.

\subsection*{\gnn{} architecture}
\textbf{Pre-Processing:} In the pre-processing layer, we generate a compact vector representation for particle and their interactions $e_{ij}$ by employing Multi-Layer Perceptrons. 
\begin{alignat}{2}
    \ch^0_i &= \sqp(\MLP_{em}(\texttt{one-hot}(t_i))) \label{eq:one-hot}\\
    \ch^0_{ij} &= \sqp(\MLP_{em}(e_{ij}))
\end{alignat}
Here, $\sqp$ is an activation function. In our implementation, we use different $\texttt{MLP}_{em}$s for node representation corresponding to kinetic energy, potential energy, and drag. For brevity, we do not separately write the $\texttt{MLP}_{em}$s in Eq.~\ref{eq:one-hot}.\\

\textbf{Kinetic energy and drag prediction.} Given that the graph employs Cartesian coordinates, the mass matrix can be represented as a diagonal matrix. Consequently, the kinetic energy ($\tau_i$) of a particle relies exclusively on the velocity ($\dot{\cx}_i$) and mass ($m_{i}$) of said particle. In this context, the parameterized masses for each particle type are acquired through the utilization of the embedding ($\ch^0_i$). As such, the predicted value of $\tau_i$ for a given particle is determined by $\tau_i = \sqp(\texttt{MLP}_{T}(\ch^0_i \parallel \dot{\x}_i))$, where the symbol $\parallel$ denotes the concatenation operator. In this equation, $\texttt{MLP}_{T}$ denotes a multilayer perceptron responsible for learning the kinetic energy function, while $\sqp$ represents the activation function employed. The overall kinetic energy of the system, denoted by $T$, is calculated as the sum of individual kinetic energies: $T=\sum_{i=1}^{n} \tau_i$.

\textbf{Potential energy prediction.} Typically, the potential energy of a system exhibits significant dependence on the topology of its underlying structure. In order to effectively capture this information, we utilize a multiple layers of message-passing among interacting particles (nodes). During the $l^{th}$ layer of message passing, the node embeddings are iteratively updated according to the following expression:
\begin{equation}
    \ch_i^{l+1} = \sqp\left(  \MLP\left(\ch_i^{l}+\sum_{j \in \mathcal{N}_i}\cW_{\CV}^l\cdot\left(\ch_j^l || \ch_{ij}^l\right) \right)\right)
\end{equation}
where, $\mathcal{N}_i=\{u_j \in\CV \mid (u_i,u_j)\in\CE \}$ is the set of neighbors of particle $u_i$. $\cW_{\CV}^{l}$ is a layer-specific learnable weight matrix.
$\ch_{ij}^l$ represents the embedding of incoming edge $e_{ij}$ on $u_i$ in the $l^{th}$ layer, which is computed as follows.
\begin{equation}
    \ch_{ij}^{l+1} = \sqp\left( \MLP\left(\ch_{ij}^{l} + \cW_{\CE}^{l}\cdot\left(\ch_i^l || \ch_{j}^l\right)\right) \right)
\end{equation}
Similar to $\cW_{\CV}^{l}$, $\cW_{\CE}^{l}$ is a layer-specific learnable weight matrix specific to the edge set. The message passing is performed over $L$ layers, where $L$ is a hyper-parameter. The final node and edge representations in the $L^{th}$ layer are denoted as $\cz_i=\ch_i^L$ and $\cz_{ij}=\ch_{ij}^L$ respectively.

The total potential energy of an $n$-body system is represented as $V= \sum_{i} v_i + \sum_{ij} v_{ij}$. Here, $v_i$ denotes the energy associated with the position of particle $i$, while $v_{ij}$ represents the energy arising from the interaction between particles $i$ and $j$. For instance, $v_i$ corresponds to the potential energy of a bob in a double pendulum, considering its position within a gravitational field. On the other hand, $v_{ij}$ signifies the energy associated with the expansion and contraction of a spring connecting two particles. In the proposed framework, the prediction for $v_i$ is given by $v_i = \sqp(\MLP_{v_i}(\ch_i^0 \parallel \x_i))$. Similarly, the prediction for the pair-wise interaction energy $v_{ij}$ is determined by $v_{ij} = \sqp(\MLP_{v_{ij}}(\cz_{ij}))$. 

The parameters of the model are trained end-to-end using the MSE loss discussed in Eq.~\ref{eq:lossfunction}.
\subsection*{Model architecture and training setup} 
\label{app:arch}
For \hgnn{}, all the MLPs are two layers deep. A square plus activation function is used for all the MLPs. 
We used 10000 data points from 100 trajectories divided into 75:25 (train: validation) to train all the models. The timestep used for the forward simulation of the pendulum system is $10^{-5}s$, for the spring and gravitational system is $10^{-3}s$, and for the LJ system is 0.0001 LJ units. All the equations of motion are integrated with the velocity-Verlet integrator. Detailed training procedures and hyper-parameters are provided in the Supplementary material. All models were trained until the decrease in loss saturates to less than 0.001 over 100 epochs. The model performance is evaluated on a forward trajectory, a task it was not explicitly trained for, of $10s$ in the case of the pendulum and $20s$ in the case of spring. Note that this trajectory is ~2-3 orders of magnitude larger than the training trajectories from which the data has been sampled. The dynamics of $n$-body system are known to be chaotic for $n \geq 2$. Hence, all the results are averaged over trajectories generated from 100 different initial conditions. 
\subsection*{Symbolic regression}
SR refers to an approach to search for equations that fit the data points and fit them rather than a parametric approach where an equation is chosen apriori to fit the data. Here, we employ the PySR package to perform the SR~\cite{cranmer2023interpretable}. PySR employs a tree-based approach for fitting the governing equation based on the operations and variables provided. Since the parametric space available for SR can be too large with every additional operation, it is important to carefully provide the minimum required input features and the operations while providing meaningful constraints on the search space. 

In the present work, we choose the addition and multiplication operation. Further, we allow polynomial fit based on a set containing (square, cube, pow(n)) operations, where pow(n) refers to power from four to ten. The loss function to fit the SR is based on the mean squared error between the predicted equation and the data points obtained from \hgnn{}. Further, the equations are selected based on a score $S$ that balances complexity $C$ and loss $L$. Specifically, the score is defined as $S = \frac{dL}{dC}$, that is, the gradient of the loss with respect to complexity. For each set of hyperparameters, we select the top 10 equations based on the scores. Further, the equation having the best score among these equations is chosen as the optimal equation. All the hyperparameters associated with the SR and the corresponding equations obtained are included in the Supplementary material.

\subsection*{Simulation environment}
All the simulations and training were carried out in the JAX environment~\cite{schoenholz2020jax,bradbury2020jax}. The graph architecture was developed using the jraph package~\cite{jraph2020github}. The experiments were conducted on a machine with Apple M1 chip having 8GB RAM and running MacOS Monterey. 
\textbf{Software packages:} numpy-1.22.1, jax-0.3.0, jax-md-0.1.20, jaxlib-0.3.0, jraph-0.0.2.dev \\
\textbf{Hardware:}
Chip: Apple M1,
Total Number of Cores: 8 (4 performance and 4 efficiency), 
Memory: 8 GB, 
System Firmware Version: 7459.101.3, 
OS Loader Version: 7459.101.3


\bibliographystyle{unsrt}
\bibliography{references}

\newpage

\appendix
\section*{\centering Supplementary Material}    


\subsection*{Comparison with baselines}
\textbf{Baselines:} In order to analyze the role of the architecture of \hgnn{} in obtaining superior performance, we consider two baselines. The first, \hnn~\citep{greydanus2019hamiltonian}, is a simple MLP that directly predicts the Hamiltonian of the system. Note that the decoupling of kinetic and potential energies is implemented in \hnn. Second, ~\hgn{}~\citep{sanchez2019hamiltonian} is a graph-based version of \hnn{}, albeit without decoupling the kinetic and potential energies. While the performance of \hnn{} has been demonstrated on several spring and pendulum systems, \hgn{}~\citep{sanchez2019hamiltonian} has been evaluated only on spring systems.\\
\textbf{Datasets and systems:} To evaluate \hgnn{}, we selected standard systems, \textit{viz}, $n$-pendulums and springs, where $n=(3,4,5)$. All the graph-based models are trained on 5-pendulum and 5-spring systems only, which are then evaluated on other system sizes. 
Further, to evaluate the zero-shot generalizability of \hgnn{} to large-scale unseen systems, we simulate 5, 50, 500-link spring systems, and 5-, 10-, and 50-link pendulum systems. We also considered a hybrid spring-pendulum system unseen during training to evaluate \hgnn{} and a gravitational system. The detailed data-generation procedure is given in Methods and Supplementary Material. The timestep used for the forward simulation of the pendulum system is $10^{-5}s$ with the data collected every 1000 timesteps, and for the spring system is $10^{-3}s$ with the data collected every 100 timesteps. Model architecture and training details are provided in Methods and Supplementary Material.\\
\textbf{Evaluation Metric:} Following the work of~\citep{lnn1}, we evaluate  performance by computing the following three error metrics, namely, (1) \textit{momentum error, (ME(t))}, (2) \textit{Energy violation error (EE(t))} given by
$ME(t)=\frac{||\hat{\mathcal{M}}(t)-\mathcal{M}(t)||_2}{||\hat{\mathcal{M}}(t)||_2+||\mathcal{M}(t)||_2}, \quad
EE(t)=\frac{||\hat{\mathcal{H}}-\mathcal{H}||_2}{(||\hat{\mathcal{H}}||_2+||\mathcal{H}||_2}$.
Note that all the variables with a hat, for example, $\hat{x}$, represent the predicted values based on the trained model, and the variables without a hat, that is $x$, represent the ground truth.
\vspace{-0.10in}
\subsection*{Energy and momentum errors of \hgnn{}}
\begin{figure}[htp]
    \centering
    \begin{subfigure}
        \centering
        \includegraphics[width=0.49\columnwidth]{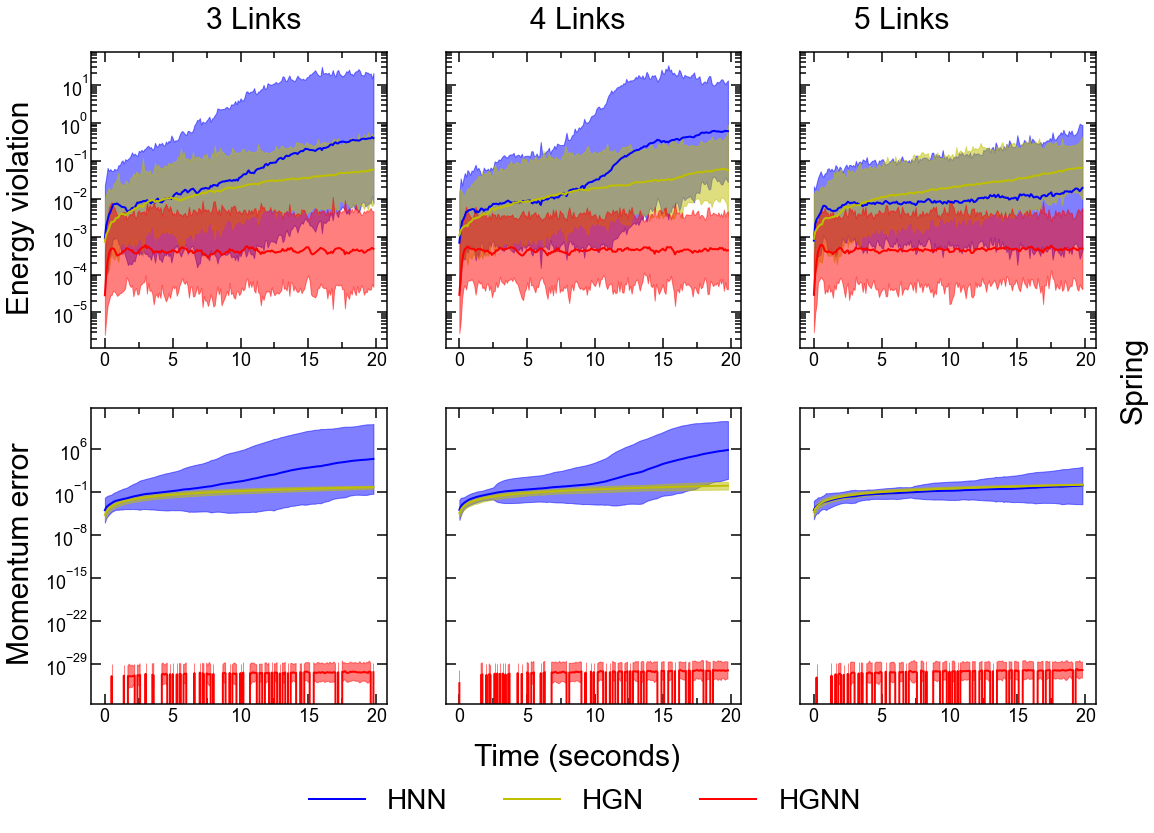}
    \end{subfigure}
    \begin{subfigure}
        \centering
\includegraphics[width=0.49\columnwidth]{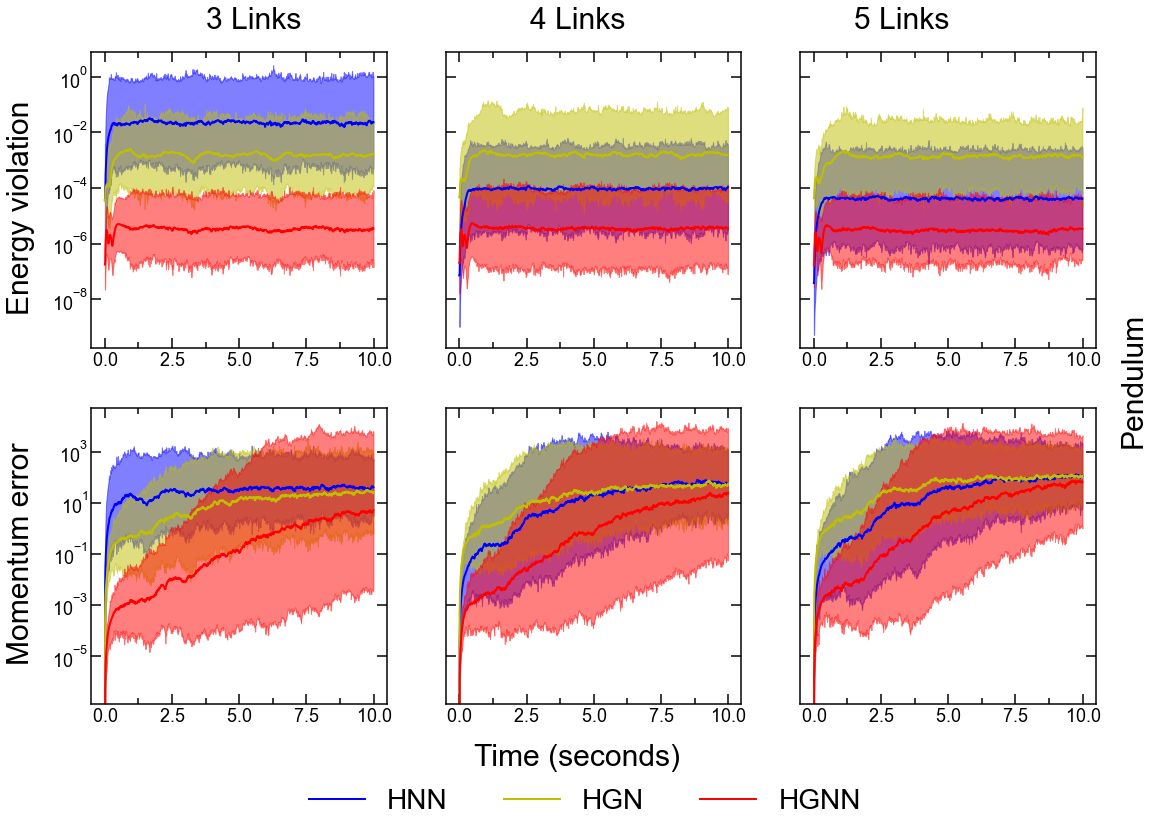}
    \end{subfigure}
    \caption{$EE$ and $PE$ for 3-,4-,5- links spring and 3-,4-,5- links pendulum systems.}
    \label{fig:ee_re_fig}
\end{figure}

\vspace{-0.10in}
Here, we analyze the evolution of energy and momentum of the trajectory predicted by \hgnn{}. We observe in the figures that the energy violation error by the \hgnn{} remains stationary and does not explode even for $10^4$ timesteps for spring and $10^5$ timesteps for the pendulum systems (see Fig.~\ref{fig:ee_re_fig}). Similarly, for the spring system, we observe that the momentum error is close to zero confirming that the total force of the system remains zero. 

\subsection*{Forces on the hybrid system}
Fig.~\ref{fig:hybrid_force_time} shows the predicted and actual force on the trajectory of all the particles for a trajectory of 100 timesteps. We observe that the predicted force is in excellent agreement with the actual force.
\begin{figure}[h]
\centering
  \includegraphics[width=0.5\linewidth]{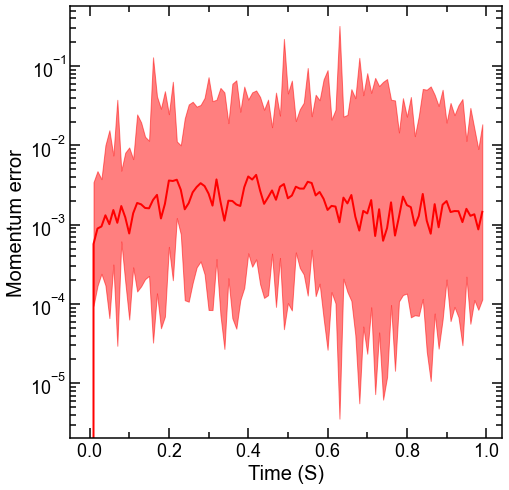}
  \caption{Time evolution of momentum error on the hybrid system of 6 particles.}
  \label{fig:mom_error}
  \vspace{-0.10in}
\end{figure}

\vspace{-0.10in}
\begin{figure}
\centering
  \includegraphics[width=0.5\linewidth]{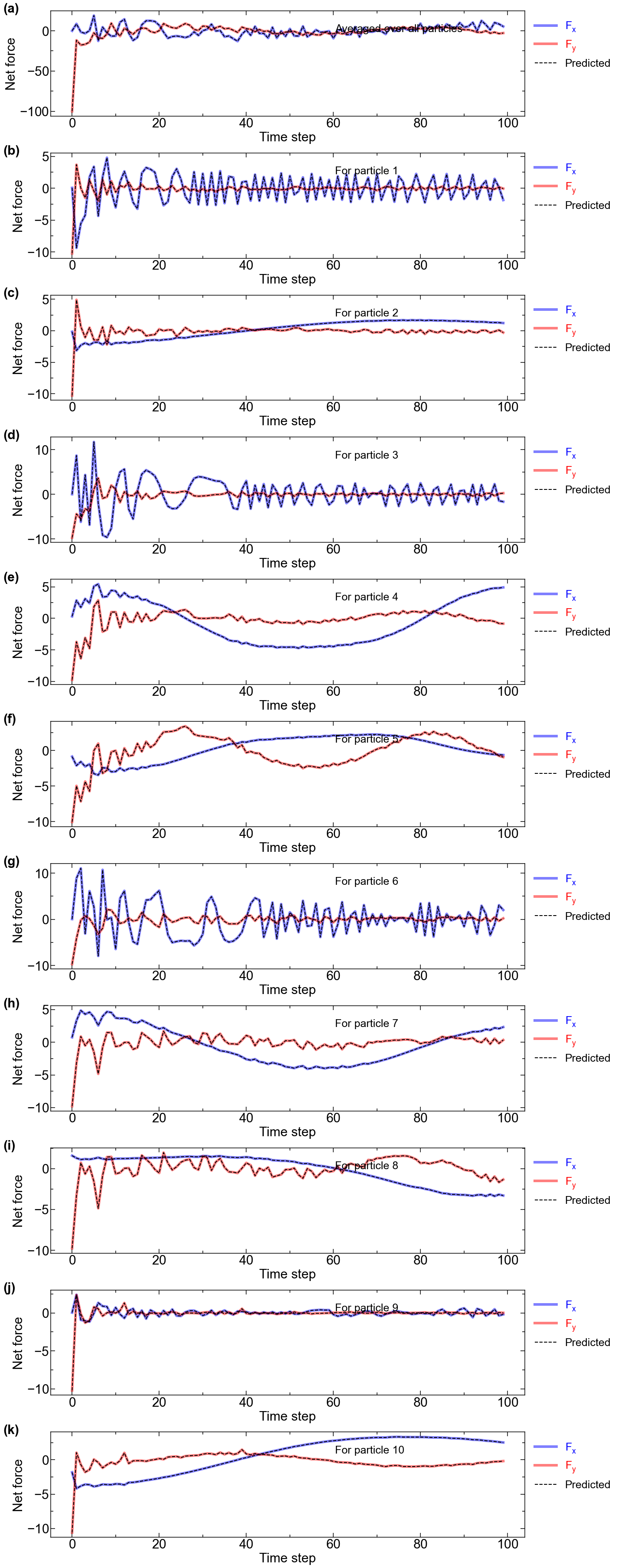}
  \caption{Time evolution of force for hybrid system on all 6 particles.}
  \label{fig:hybrid_force_time}
  \vspace{-0.10in}
\end{figure}
Fig.~\ref{fig:ee_re_fig} shows the performance of \hnn{}, \hgn{}, and \hgnn{} for spring and pendulum systems. We observe that \hgnn{} outperforms both \hnn{} and \hgn{} on both spring and pendulum systems. Specifically, we observe that the energy violation error in \hgnn{} remains saturated, suggesting a stable and realistic predicted trajectory. Note that \hnn{} is trained and evaluated on each of these systems separately, while \hgn{} and \hgnn{} are trained in only one system and inferred for all other systems by performing the forward simulation.
\vspace{-0.10in}
\subsection*{Complex systems}
\vspace{-0.10in}
\begin{figure}
\centering
  \includegraphics[width=\linewidth]{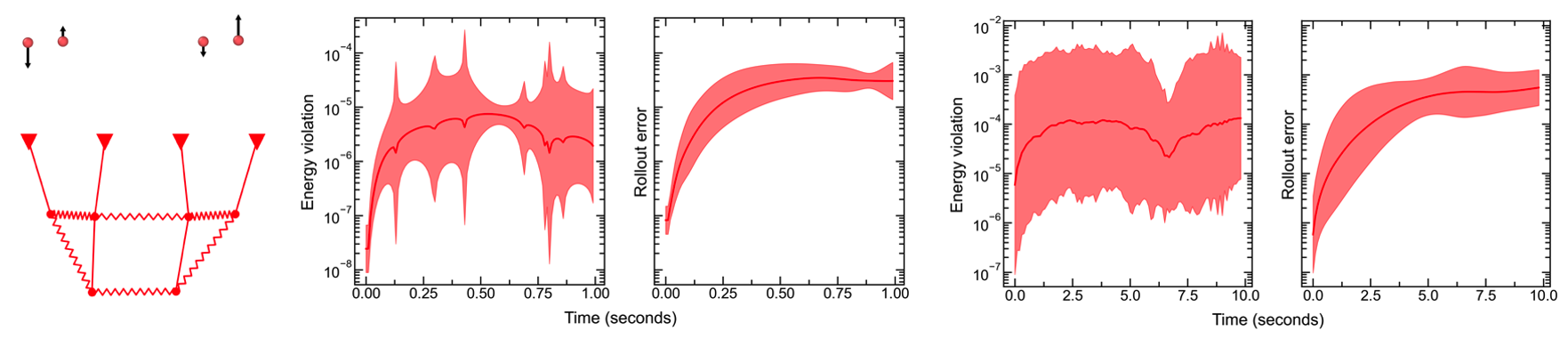}
  \vspace{-0.25in}
  \caption{Visualization of (a) gravitational and (b) hybrid systems. $EE$ and $RE$ for (c) hybrid system and (d) 4-body gravitational system.}
  \label{fig:hybrid_ee_re}
  \vspace{-0.10in}
\end{figure}

In order to evaluate the performance of \hgnn{} on more complex systems, we consider a gravitation system and a hybrid spring-pendulum system (see Figs.~\ref{fig:hybrid_ee_re}(a) and (b)). We observe that \hgnn{}, trained on spring and pendulum systems separately, provides an excellent inference for the hybrid system unseen by the model. Despite best efforts, the \hgn{} and \hnn{} was unable to provide a forward trajectory for the hybrid system. The superior performance of \hgnn{} could be attributed to the architecture, which decouples the potential and kinetic energies and learns them separately for each system. We also evaluate \hgnn{} for a more complex interaction than springs and pendulums, that is, gravitational forces. Fig.~\ref{fig:hybrid_ee_re} shows that \hgnn{} provides excellent inference for the gravitational system. Similar to the hybrid system, the baselines trained on the gravitational systems were unable to provide a stable trajectory and exploded after a few steps during the inference.
\vspace{-0.10in}
\subsection*{Zero-shot generalization}
\vspace{-0.10in}
\begin{figure}
    \centering
    \begin{subfigure}
        \centering
        \includegraphics[width=0.49\columnwidth]{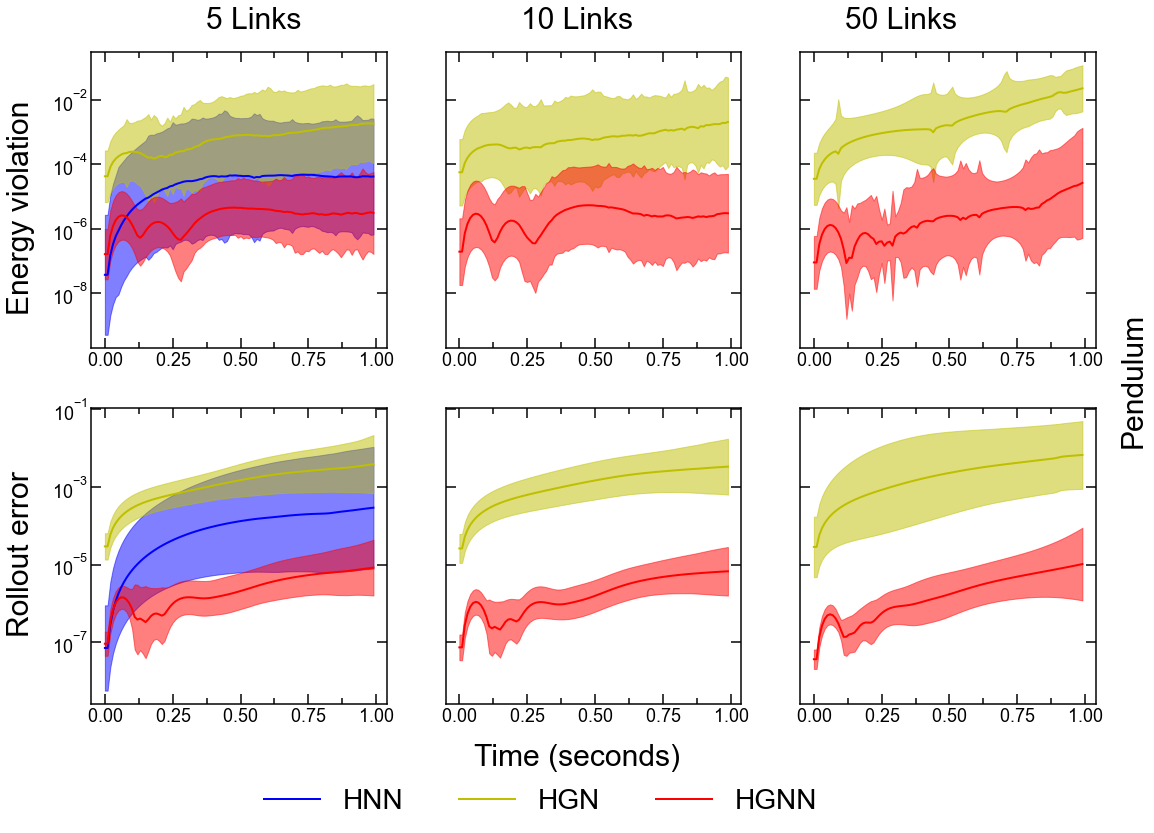}
    \end{subfigure}
    \begin{subfigure}
        \centering
        \includegraphics[width=0.49\columnwidth]{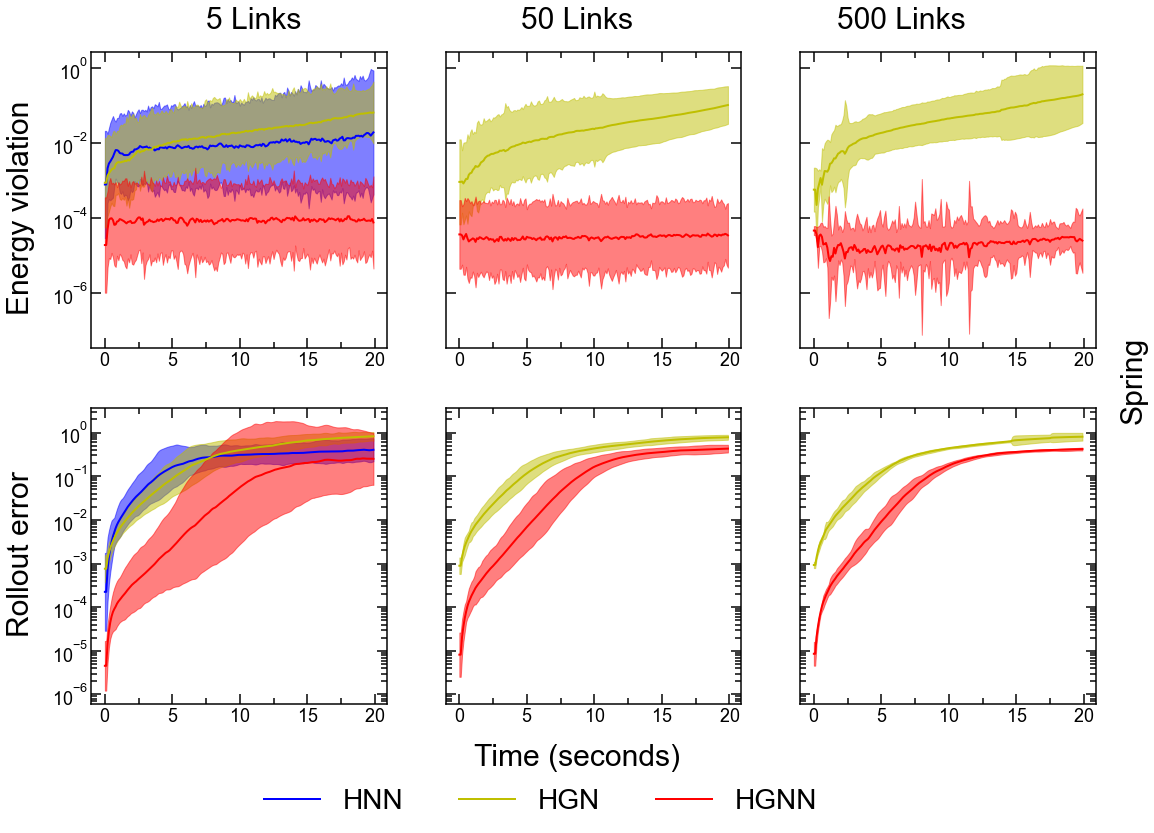}
       \vspace{-0.20in}
    \end{subfigure}
    \caption{Energy violation and Rollout error for 5-,10-,50-links pendulum and 5-,50-,500-links spring systems.}
    \label{fig:gen_fig}
   \vspace{-0.20in}
\end{figure}

Finally, we evaluate the zero-shot generalizability of \hgnn{} in comparison to \hgn{} (see Fig.~\ref{fig:gen_fig}). We observe that \hgnn{} exhibits superior generalization to system sizes that are two orders of magnitude larger than the training system. In the case of the spring system, even for a system size two orders of magnitude error, we observe a comparable error in energy, which remains stable with time.
\subsection*{Hyper-parameters}
\label{app:hyper}
The hyper-parameters used for training each of the architectures are provided below. 

$\bullet$\textbf{\hgnn}
\begin{center}
\begin{tabular}{ |c|c| } 
 \hline
 \textbf{Parameter} & \textbf{Value} \\
 \hline
 Node embedding dimension & 5 \\ 
 Edge embedding dimension & 5 \\ 
 Hidden layer neurons (MLP) & 5 \\ 
 Number of hidden layers (MLP) & 2 \\ 
 Activation function & squareplus\\
 Number of layers of message passing(pendulum) & 2\\
  Number of layers of message passing(spring) & 1\\
 Optimizer & ADAM \\
 Learning rate & $1.0e^{-3}$ \\
 Batch size & 100 \\
 \hline
\end{tabular}
\end{center}

$\bullet$\textbf{\hnn}
\begin{center}
\begin{tabular}{ |c|c| } 
 \hline
 \textbf{Parameter} & \textbf{Value} \\
 \hline
 Hidden layer neurons (MLP) & 256 \\ 
 Number of hidden layers (MLP) & 2 \\ 
 Activation function & squareplus\\
 Optimizer & ADAM \\
 Learning rate & $1.0e^{-3}$ \\
 Batch size & 100 \\
 \hline
\end{tabular}
\end{center}

$\bullet$\textbf{\hgn}
\begin{center}
\begin{tabular}{ |c|c| } 
 \hline
 \textbf{Parameter} & \textbf{Value} \\
 \hline
 Node embedding dimension & 8 \\ 
 Edge embedding dimension & 8 \\ 
 Hidden layer neurons (MLP) & 16 \\ 
 Number of hidden layers (MLP) & 2 \\ 
 Activation function & squareplus\\
 Number of layers of message passing & 1\\
 Optimizer & ADAM \\
 Learning rate & $1.0e^{-3}$ \\
 Batch size & 100 \\
 \hline
\end{tabular}
\end{center}

\subsection*{Symbolic Regression}
The equations obtained for each of the systems from the symbolic regression, the loss and the scores are provided in this section.

$\bullet$\textbf{Kinetic Energy vs velocity}
\begin{table}[h]
\begin{center}
\begin{tabular}{@{}cccc@{}}
\toprule
Equation & Complexity & Loss & Score \\
\midrule
$y = x_{0}$ & $1$ & $15.2$ & $0.0$ \\
$y = 2.12 x_{0}$ & $3$ & $5.69$ & $0.489$ \\
$\mathbf{y = 0.500 x_{0}^{2}}$ & $\mathbf{4}$ & $\mathbf{7.96 \cdot 10^{-10}}$ & $\mathbf{22.7}$ \\
$y = 0.500 x_{0}^{2} + 1.87 \cdot 10^{-5}$ & $8$ & $5.74 \cdot 10^{-10}$ & $0.0816$ \\
$y = 0.500 x_{0}^{2} + 1.81 \cdot 10^{-5}$ & $19$ & $5.71 \cdot 10^{-10}$ & $0.000542$ \\
$y = 0.500 x_{0}^{2} + 1.76 \cdot 10^{-5}$ & $20$ & $5.63 \cdot 10^{-10}$ & $0.0145$ \\
\bottomrule
\end{tabular}
\end{center}
\end{table}

$\bullet$\textbf{Spring potential energy vs position}
\begin{table}[h]
\begin{center}
\begin{tabular}{@{}cccc@{}}
\toprule
Equation & Complexity & Loss & Score \\
\midrule
$y = 0.00590$ & $1$ & $1.72 \cdot 10^{-5}$ & $0.0$ \\
$y = 0.489 \left(x_{0} - 1.00\right)^{2}$ & $6$ & $2.65 \cdot 10^{-8}$ & $1.30$ \\
$\mathbf{y = 0.499 \left(x_{0} - 1.00\right)^{2}}$ & $\mathbf{7}$ & $\mathbf{1.13 \cdot 10^{-9}}$ & $\mathbf{3.15}$ \\
$y = 0.500 \left(x_{0} - 1.00\right)^{2}$ & $8$ & $4.12 \cdot 10^{-10}$ & $1.01$ \\
\bottomrule
\end{tabular}
\end{center}
\end{table}




$\bullet$\textbf{Pair-wise LJ interactions}

Pairwise LJ interactions are obtained by conducting a parametric study with different polynomial orders. Tables 2, 3, and 4 show the best equations obtained for (0-0), (0-1), and (1-1) interactions. The polynomials considered corresponding to each equation are provided as Power in the table. The detailed results obtained for each combination of polynomials are included in the following tables.

\begin{table}[h]
\caption{LJ potential for 0-0 interaction} 
\begin{center}    
\begin{tabular}{@{}ccccc@{}}
\toprule
Equation & Complexity & Loss & Score & Power \\
\midrule
$\begin{aligned} y &= 7.92 \left(0.502 x_{0} - 1\right)^{3} + \frac{1}{x_{0}^{15}} \end{aligned}$ & $11$ & $0.00114$ & $1.93$ & $2, 3$ \\
\\
$\begin{aligned} y &= \frac{-0.857 + \frac{0.871}{x_{0}^{12}}}{x_{0}^{4}} \end{aligned}$ & $10$ & $0.00181$ & $2.48$ & $2, 3, 4$ \\
\\
$\begin{aligned} y &= - \frac{1.03}{x_{0}^{5}} + \frac{1.05}{x_{0}^{15}} \end{aligned}$ & $10$ & $0.000890$ & $1.94$ & $2, 3, 5$ \\
\\
$\mathbf{y = \frac{-1.95 + \frac{1.90}{x_{0}^{6}}}{x_{0}^{6}}}$ & $\mathbf{9}$ & $\mathbf{0.00159}$ & $\mathbf{2.62}$ & $\mathbf{2, 3, 6}$ \\
\\
$y = \frac{-1.62 + \frac{1.64}{x_{0}^{7}}}{x_{0}^{6}}$ & $11$ & $0.000187$ & $1.94$ & $2, 3, 7$ \\
\\
$\begin{aligned} y &= \frac{1.65 \left(-0.986 + \frac{1}{x_{0}^{7}}\right)}{x_{0}^{6}} \end{aligned}$ & $12$ & $0.000187$ & $2.44$ & $2, 3, 8$ \\
\\
$y = \frac{-1.21 + \frac{1.19}{x_{0}^{9}}}{x_{0}^{6}}$ & $10$ & $0.000926$ & $2.20$ & $2, 3, 9$ \\
\\
$y = \frac{-1.14 + \frac{1.09}{x_{0}^{10}}}{x_{0}^{6}}$ & $10$ & $0.00145$ & $1.96$ & $2, 3, 10$ \\
\\
\bottomrule
\end{tabular}
\end{center}
\end{table}

\begin{table}[h]
\caption{LJ potential for 0-1 interaction}
\begin{center}
\begin{tabular}{@{}ccccc@{}}
\toprule
Equation & Complexity & Loss & Score & Power\\
\\
\midrule
$y = \frac{-2.67 + \frac{2.14}{x_{0}}}{x_{0}^{9}}$ & $9$ & $0.000697$ & $3.68$ & $2, 3$ \\
\\
$y = \frac{-2.91 + \frac{2.33}{x_{0}}}{x_{0}^{8}}$ & $9$ & $3.47 \cdot 10^{-5}$ & $3.32$ & $2, 3, 4$ \\
\\
$y = \frac{-2.91 + \frac{2.33}{x_{0}}}{x_{0}^{8}}$ & $10$ & $3.47 \cdot 10^{-5}$ & $3.00$ & $2, 3, 5$ \\
\\
$y = \frac{2.33 - 2.91 x_{0}}{x_{0}^{9}}$ & $9$ & $3.47 \cdot 10^{-5}$ & $5.98$ & $2, 3, 6$ \\
\\
$\begin{aligned} y &= \frac{-0.764 + \frac{0.154}{x_{0}^{7}}}{x_{0}^{6}} \end{aligned}$ & $10$ & $0.00149$ & $1.18$ & $2, 3, 7$ \\
\\
$y = \frac{-2.91 + \frac{2.33}{x_{0}}}{x_{0}^{8}}$ & $8$ & $3.47 \cdot 10^{-5}$ & $6.70$ & $2, 3, 8$ \\
\\
$y = \frac{-2.67 + \frac{2.14}{x_{0}}}{x_{0}^{9}}$ & $8$ & $0.000697$ & $3.68$ & $2, 3, 9$ \\
\\
$\begin{aligned} y &= \frac{-0.773 + \frac{0.203}{x_{0}^{6}}}{x_{0}^{6}} \end{aligned}$ & $11$ & $0.000109$ & $2.22$ & $2, 3, 10$ \\
\\
\bottomrule
\end{tabular}
\end{center}
\end{table}

\begin{table}[h]
\caption{LJ potential for 1-1 interaction}
\begin{center}    
\begin{tabular}{@{}ccccc@{}}
\toprule
Equation & Complexity & Loss & Score & power\\
\midrule
$\mathbf{y = \frac{-0.464 + \frac{0.215}{x_{0}^{6}}}{x_{0}^{6}}}$ & $\mathbf{11}$ & $\mathbf{1.16 \cdot 10^{-5}}$ & $\mathbf{5.41}$ & $\mathbf{2, 3}$ \\
\\
$\begin{aligned} y &= \frac{- \frac{0.462}{x_{0}^{2}} + \frac{0.215}{x_{0}^{8}}}{x_{0}^{4}} \end{aligned}$ & $13$ & $1.17 \cdot 10^{-5}$ & $2.78$ & $2, 3, 4$ \\
\\
$y = \frac{-1.97 + \frac{1.73}{x_{0}}}{x_{0}^{9}}$ & $9$ & $0.000136$ & $3.30$ & $2, 3, 5$ \\
\\
$\begin{aligned} y &= 0.00379 + \frac{-0.471 + \frac{0.217}{x_{0}^{6}}}{x_{0}^{6}} \end{aligned}$ & $11$ & $5.75 \cdot 10^{-7}$ & $3.00$ & $2, 3, 6$ \\
\\
$\begin{aligned} y &= \frac{-0.642 + \frac{0.139}{x_{0}^{12}}}{x_{0}^{7} + x_{0}} \end{aligned}$ & $13$ & $1.94 \cdot 10^{-6}$ & $5.01$ & $2, 3, 7$ \\
\\
$\begin{aligned} y &= \frac{-0.463 + \frac{0.215}{x_{0}^{6}}}{x_{0}^{6}} \end{aligned}$ & $11$ & $1.16 \cdot 10^{-5}$ & $4.59$ & $2, 3, 8$ \\
\\
$\begin{aligned} y &= \frac{- \frac{0.376}{x_{0}} + \frac{0.138}{x_{0}^{9}}}{x_{0}^{4}} \end{aligned}$ & $12$ & $0.000108$ & $1.68$ & $2, 3, 9$ \\
\\
$\begin{aligned} y &= \frac{-0.289 + \frac{0.0700}{x_{0}^{10}}}{x_{0}^{6}} \end{aligned}$ & $10$ & $0.000322$ & $1.79$ & $2, 3, 10$ \\
\\
\bottomrule
\end{tabular}
\end{center}
\end{table}

\begin{table}[h]
\caption{LJ potential for 0-0 interaction || Power 2,3}
\begin{center}
\begin{tabular}{@{}cccc@{}}
\toprule
Equation & Complexity & Loss & Score \\
\\
\midrule
$y = -0.0554$ & $1$ & $0.0688$ & $0.0$ \\
\\
$y = -0.0553$ & $2$ & $0.0688$ & $7.27 \cdot 10^{-8}$ \\
\\
$y = - \frac{0.0952}{x_{0}}$ & $3$ & $0.0683$ & $0.00693$ \\
\\
\bottomrule
\end{tabular}
\end{center}
\end{table}

\begin{table}[h]
\caption{LJ potential for 0-0 interaction || Power 2,3,5}
\begin{center}
\begin{tabular}{@{}cccc@{}}
\toprule
Equation & Complexity & Loss & Score \\
\\
\midrule
$y = -0.0554$ & $1$ & $0.0688$ & $0.0$ \\
\\
$y = -0.0553$ & $2$ & $0.0688$ & $2.18 \cdot 10^{-7}$ \\
\\
$y = - \frac{0.0953}{x_{0}}$ & $3$ & $0.0683$ & $0.00693$ \\
\\
$\begin{aligned} y &= 4.04 \cdot 10^{4} \left(0.515 x_{0} - 1\right)^{16} \end{aligned}$ & $5$ & $0.0423$ & $0.239$ \\
\\
$\begin{aligned} y &= 1.64 \cdot 10^{9} \left(0.515 x_{0} - 1\right)^{32} \end{aligned}$ & $6$ & $0.0279$ & $0.417$ \\
\\
$\begin{aligned} y &= 1.68 \cdot 10^{9} \left(0.515 x_{0} - 1\right)^{32} - 0.0854 \end{aligned}$ & $8$ & $0.0209$ & $0.144$ \\
\\
$\begin{aligned} y &= 7.97 \left(0.501 x_{0} - 1\right)^{3} + \frac{1}{x_{0}^{15}} \end{aligned}$ & $10$ & $0.00116$ & $1.45$ \\
\\
$\begin{aligned} y &= 7.97 \left(0.501 x_{0} - 1\right)^{3} + \frac{1}{x_{0}^{15}} \end{aligned}$ & $11$ & $0.00116$ & $2.58 \cdot 10^{-7}$ \\
\\
$\begin{aligned} y &= \frac{x_{0} - 2.01 + \frac{1.03}{x_{0}^{12}}}{x_{0}^{3}} \end{aligned}$ & $12$ & $0.000354$ & $1.19$ \\
\\
$\begin{aligned} y &= \frac{x_{0} - 2.01 + \frac{1.03}{x_{0}^{12}}}{x_{0}^{3}} \end{aligned}$ & $13$ & $0.000354$ & $5.65 \cdot 10^{-7}$ \\
\\
$\begin{aligned} y &= \frac{x_{0} - 2.55 + \frac{1.57}{x_{0}^{8}}}{x_{0}^{5}} \end{aligned}$ & $14$ & $5.81 \cdot 10^{-5}$ & $1.81$ \\
\\
$\begin{aligned} y &= \frac{x_{0} - 2.55 + \frac{1.57}{x_{0}^{8}}}{x_{0}^{5}} \end{aligned}$ & $16$ & $5.81 \cdot 10^{-5}$ & $1.63 \cdot 10^{-7}$ \\
\\
$\begin{aligned} y &= \frac{x_{0} - 2.55 + \frac{1.57}{x_{0}^{8}}}{x_{0}^{5}} \end{aligned}$ & $17$ & $5.80 \cdot 10^{-5}$ & $0.00138$ \\
\\
$\begin{aligned} y &= \frac{x_{0} \left(x_{0} - 2.56 + \frac{1.59}{x_{0}^{8}}\right)}{x_{0}^{6} + 0.00348} \end{aligned}$ & $18$ & $5.70 \cdot 10^{-5}$ & $0.0172$ \\
\\
$\begin{aligned} y &= \frac{x_{0} \left(x_{0} - 2.56 + \frac{1.59}{x_{0}^{8}}\right)}{x_{0}^{6} + 0.00837} \end{aligned}$ & $19$ & $5.42 \cdot 10^{-5}$ & $0.0509$ \\
\\
\bottomrule
\end{tabular}
\end{center}
\end{table}

\begin{table}[h]
\caption{LJ potential for 0-0 interaction || Power 2,3,4}
\begin{center}
\begin{tabular}{@{}cccc@{}}
\toprule
Equation & Complexity & Loss & Score \\
\\
\midrule
$y = -0.0551$ & $1$ & $0.0688$ & $0.0$ \\
\\
$y = - \frac{0.0952}{x_{0}}$ & $3$ & $0.0683$ & $0.00347$ \\
\\
$y = \frac{0.235}{x_{0}^{25}}$ & $5$ & $0.0299$ & $0.414$ \\
\\
$y = \frac{0.152}{x_{0}^{30}}$ & $6$ & $0.0290$ & $0.0296$ \\
\\
$y = -0.0975 + \frac{0.256}{x_{0}^{25}}$ & $7$ & $0.0218$ & $0.286$ \\
\\
$y = \frac{-1.31 + \frac{1.36}{x_{0}^{5}}}{x_{0}^{5}}$ & $9$ & $0.0135$ & $0.239$ \\
\\
$y = \frac{-1.10 + \frac{1.11}{x_{0}^{10}}}{x_{0}^{5}}$ & $10$ & $0.000653$ & $3.03$ \\
\\
$\begin{aligned} y &= - \frac{1.87}{x_{0}^{6}} + \frac{1.92}{x_{0}^{12}} \end{aligned}$ & $12$ & $0.000323$ & $0.352$ \\
\\
$\begin{aligned} y &= - \frac{1.87}{x_{0}^{6}} + \frac{1.92}{x_{0}^{12}} \end{aligned}$ & $13$ & $0.000323$ & $3.10 \cdot 10^{-7}$ \\
\\
$\begin{aligned} y &= 0.0128 - \frac{1.87}{x_{0}^{6}} + \frac{1.92}{x_{0}^{12}} \end{aligned}$ & $14$ & $0.000140$ & $0.833$ \\
\\
$\begin{aligned} y &= 0.0128 - \frac{1.87}{x_{0}^{6}} + \frac{1.92}{x_{0}^{12}} \end{aligned}$ & $15$ & $0.000140$ & $3.56 \cdot 10^{-7}$ \\
\\
$\begin{aligned} y &= - \frac{3.53}{x_{0}^{8} + 2.71} + \frac{1}{x_{0}^{14}} \end{aligned}$ & $17$ & $6.21 \cdot 10^{-5}$ & $0.408$ \\
\\
$\begin{aligned} y &= - \frac{3.53}{x_{0}^{8} + x_{0} + 1.95} + \frac{0.922}{x_{0}^{15}} \end{aligned}$ & $18$ & $2.03 \cdot 10^{-5}$ & $1.12$ \\
\\
$\begin{aligned} y &= - \frac{3.53}{x_{0}^{8} + x_{0} + 1.98} + \frac{0.922}{x_{0}^{15}} \end{aligned}$ & $20$ & $1.60 \cdot 10^{-5}$ & $0.118$ \\
\\
\bottomrule
\end{tabular}
\end{center}
\end{table}

\begin{table}[h]
\caption{LJ potential for 0-0 interaction || Power 2, 3, 6}
\begin{center}
\begin{tabular}{@{}cccc@{}}
\toprule
Equation & Complexity & Loss & Score \\
\\
\midrule
$y = -0.0552$ & $1$ & $0.0688$ & $0.0$ \\
\\
$y = -0.0554$ & $2$ & $0.0688$ & $7.27 \cdot 10^{-7}$ \\
\\
$y = - \frac{0.0952}{x_{0}}$ & $3$ & $0.0683$ & $0.00693$ \\
\\
\bottomrule
\end{tabular}
\end{center}
\end{table}

\begin{table}[h]
\caption{LJ potential for 0-0 interaction || Power 2, 3, 7}
\begin{center}
\begin{tabular}{@{}cccc@{}}
\toprule
Equation & Complexity & Loss & Score \\
\\
\midrule
$y = -0.0554$ & $1$ & $0.0688$ & $0.0$ \\
\\
$y = - \frac{0.0952}{x_{0}}$ & $3$ & $0.0683$ & $0.00347$ \\
\\
$y = \frac{0.310}{x_{0}^{21}}$ & $5$ & $0.0336$ & $0.355$ \\
\\
$y = \frac{0.183}{x_{0}^{28}}$ & $6$ & $0.0290$ & $0.149$ \\
\\
$y = -0.0966 + \frac{0.362}{x_{0}^{21}}$ & $7$ & $0.0247$ & $0.160$ \\
\\
$y = \frac{-11.5 + \frac{11.6}{x_{0}}}{x_{0}^{7}}$ & $8$ & $0.00386$ & $1.86$ \\
\\
$y = \frac{-11.9 + \frac{11.9}{x_{0}}}{x_{0}^{9}}$ & $9$ & $0.000160$ & $3.18$ \\
\\
$y = \frac{-2.88 + \frac{2.93}{x_{0}^{4}}}{x_{0}^{7}}$ & $10$ & $8.96 \cdot 10^{-5}$ & $0.579$ \\
\\
$\begin{aligned} y &= \frac{- \frac{5.85}{x_{0}} + \frac{5.90}{x_{0}^{3}}}{x_{0}^{7}} \end{aligned}$ & $11$ & $2.91 \cdot 10^{-5}$ & $1.12$ \\
\\
$y = \frac{-5.85 + \frac{5.90}{x_{0}^{2}}}{x_{0}^{8}}$ & $12$ & $2.91 \cdot 10^{-5}$ & $5.84 \cdot 10^{-7}$ \\
\\
$\begin{aligned} y &= 0.00348 + \frac{-5.85 + \frac{5.90}{x_{0}^{2}}}{x_{0}^{8}} \end{aligned}$ & $13$ & $1.37 \cdot 10^{-5}$ & $0.753$ \\
\\
$\begin{aligned} y &= 0.00348 + \frac{-5.85 + \frac{5.90}{x_{0}^{2}}}{x_{0}^{8}} \end{aligned}$ & $14$ & $1.37 \cdot 10^{-5}$ & $8.02 \cdot 10^{-6}$ \\
\\
$\begin{aligned} y &= 0.00406 + \frac{-5.85 + \frac{5.90}{x_{0}^{2}}}{x_{0}^{8}} \end{aligned}$ & $15$ & $1.35 \cdot 10^{-5}$ & $0.0159$ \\
\\
$\begin{aligned} y &= 0.00388 + \frac{-5.85 + \frac{5.90}{x_{0}^{2}}}{x_{0}^{8}} \end{aligned}$ & $16$ & $1.35 \cdot 10^{-5}$ & $0.000387$ \\
\\
$\begin{aligned} y &= 0.00534 + \frac{-5.90 + \frac{5.94}{x_{0}^{2}}}{x_{0}^{8}} \end{aligned}$ & $17$ & $8.02 \cdot 10^{-6}$ & $0.521$ \\
\\
$\begin{aligned} y &= 0.00534 + \frac{-5.90 + \frac{5.94}{x_{0}^{2}}}{x_{0}^{8}} \end{aligned}$ & $20$ & $8.02 \cdot 10^{-6}$ & $1.21 \cdot 10^{-6}$ \\
\\
\bottomrule
\end{tabular}
\end{center}
\end{table}

\begin{table}[h]
\caption{LJ potential for 0-0 interaction || Power 2, 3, 8}
\begin{center}
\begin{tabular}{@{}cccc@{}}
\toprule
Equation & Complexity & Loss & Score \\
\\
\midrule
$y = -0.0553$ & $1$ & $0.0688$ & $0.0$ \\
\\
$y = - \frac{0.0952}{x_{0}}$ & $3$ & $0.0683$ & $0.00347$ \\
\\
$y = 129. \left(0.545 x_{0} - 1\right)^{8}$ & $4$ & $0.0623$ & $0.0932$ \\
\\
$y = \frac{0.254}{x_{0}^{24}}$ & $5$ & $0.0305$ & $0.713$ \\
\\
$y = \frac{0.202}{x_{0}^{27}}$ & $6$ & $0.0291$ & $0.0465$ \\
\\
$y = -0.0914 + \frac{0.262}{x_{0}^{24}}$ & $7$ & $0.0222$ & $0.271$ \\
\\
$y = \frac{-11.5 + \frac{11.6}{x_{0}}}{x_{0}^{8}}$ & $8$ & $0.000361$ & $4.12$ \\
\\
$y = \frac{-5.85 + \frac{5.90}{x_{0}^{2}}}{x_{0}^{8}}$ & $9$ & $2.91 \cdot 10^{-5}$ & $2.52$ \\
\\
$y = \frac{-5.85 + \frac{5.90}{x_{0}^{2}}}{x_{0}^{8}}$ & $10$ & $2.91 \cdot 10^{-5}$ & $1.24 \cdot 10^{-6}$ \\
\\
$\begin{aligned} y &= 0.00348 + \frac{-5.85 + \frac{5.90}{x_{0}^{2}}}{x_{0}^{8}} \end{aligned}$ & $11$ & $1.37 \cdot 10^{-5}$ & $0.753$ \\
\\
$\begin{aligned} y &= 0.00348 + \frac{-5.85 + \frac{5.90}{x_{0}^{2}}}{x_{0}^{8}} \end{aligned}$ & $12$ & $1.37 \cdot 10^{-5}$ & $8.16 \cdot 10^{-6}$ \\
\\
$\begin{aligned} y &= 0.00388 + \frac{-5.85 + \frac{5.90}{x_{0}^{2}}}{x_{0}^{8}} \end{aligned}$ & $13$ & $1.35 \cdot 10^{-5}$ & $0.0164$ \\
\\
$\begin{aligned} y &= \frac{-5.85 + \frac{5.90}{x_{0}^{2}}}{0.00121 x_{0}^{16} + x_{0}^{8}} \end{aligned}$ & $15$ & $1.00 \cdot 10^{-5}$ & $0.149$ \\
\\
$\begin{aligned} y &= \frac{-5.85 + \frac{5.90}{x_{0}^{2}}}{0.000522 x_{0}^{18} + x_{0}^{8}} \end{aligned}$ & $16$ & $8.00 \cdot 10^{-6}$ & $0.224$ \\
\\
\bottomrule
\end{tabular}
\end{center}
\end{table}

\begin{table}[h]
\caption{LJ potential for 0-0 interaction || Power 2, 3, 9}
\begin{center}
\begin{tabular}{@{}cccc@{}}
\toprule
Equation & Complexity & Loss & Score \\
\\
\midrule
$y = -0.0553$ & $1$ & $0.0688$ & $0.0$ \\
\\
$y = - \frac{0.0957}{x_{0}}$ & $3$ & $0.0683$ & $0.00347$ \\
\\
$y = \frac{0.259}{x_{0}^{9}}$ & $4$ & $0.0641$ & $0.0639$ \\
\\
$y = \frac{0.200}{x_{0}^{27}}$ & $5$ & $0.0291$ & $0.789$ \\
\\
$y = -0.0885 + \frac{0.213}{x_{0}^{27}}$ & $7$ & $0.0215$ & $0.152$ \\
\\
$\begin{aligned} y &= - \frac{11.5 \left(x_{0} - 1.01\right)}{x_{0}^{9}} \end{aligned}$ & $9$ & $0.000361$ & $2.04$ \\
\\
$\begin{aligned} y &= - \frac{5.85 \left(x_{0} - \frac{1.01}{x_{0}}\right)}{x_{0}^{9}} \end{aligned}$ & $11$ & $2.91 \cdot 10^{-5}$ & $1.26$ \\
\\
$\begin{aligned} y &= 0.00442 - \frac{5.85 \left(x_{0} - \frac{1.01}{x_{0}}\right)}{x_{0}^{9}} \end{aligned}$ & $13$ & $1.38 \cdot 10^{-5}$ & $0.374$ \\
\\
$\begin{aligned} y &= 0.00373 - \frac{5.85 \left(x_{0} - \frac{1.01}{x_{0}}\right)}{x_{0}^{9}} \end{aligned}$ & $14$ & $1.36 \cdot 10^{-5}$ & $0.0133$ \\
\\
$\begin{aligned} y &= \frac{\left(x_{0} - \frac{1.01}{x_{0}}\right) \left(\left(x_{0} - 0.799\right)^{3} - 5.85\right)}{x_{0}^{9}} \end{aligned}$ & $16$ & $1.14 \cdot 10^{-5}$ & $0.0897$ \\
\\
$\begin{aligned} y &= - \frac{5.85}{\frac{x_{0}^{9}}{x_{0} - \frac{1.01}{x_{0}}} + 21.9 \left(x_{0} - 0.715\right)^{9}} \end{aligned}$ & $18$ & $9.73 \cdot 10^{-6}$ & $0.0773$ \\
\\
$\begin{aligned} y &= - \frac{5.85}{\frac{x_{0}^{9}}{x_{0} - \frac{1.01}{x_{0}}} + 21.9 \left(x_{0} - 0.757\right)^{9}} \end{aligned}$ & $19$ & $7.98 \cdot 10^{-6}$ & $0.198$ \\
\\
$\begin{aligned} y &= - \frac{5.85}{\frac{x_{0}^{9}}{x_{0} - \frac{1.01}{x_{0}}} + 21.9 \left(x_{0} - 0.757\right)^{9}} \end{aligned}$ & $20$ & $7.98 \cdot 10^{-6}$ & $2.51 \cdot 10^{-7}$ \\
\\
\bottomrule
\end{tabular}
\end{center}
\end{table}

\begin{table}[h]
\caption{LJ potential for 0-0 interaction || Power 2, 3, 10}
\begin{center}
\begin{tabular}{@{}cccc@{}}
\toprule
Equation & Complexity & Loss & Score \\
\\
\midrule
$y = -0.0553$ & $1$ & $0.0688$ & $0.0$ \\
\\
$y = - \frac{0.0952}{x_{0}}$ & $3$ & $0.0683$ & $0.00347$ \\
\\
$y = 589. \left(0.528 x_{0} - 1\right)^{10}$ & $4$ & $0.0574$ & $0.175$ \\
\\
$y = \frac{0.152}{x_{0}^{30}}$ & $5$ & $0.0290$ & $0.682$ \\
\\
$y = -0.0929 + \frac{0.165}{x_{0}^{30}}$ & $7$ & $0.0221$ & $0.136$ \\
\\
$y = \frac{-11.9 + \frac{11.8}{x_{0}}}{x_{0}^{10}}$ & $8$ & $0.00144$ & $2.73$ \\
\\
$y = \frac{-11.9 + \frac{11.9}{x_{0}}}{x_{0}^{9}}$ & $9$ & $0.000160$ & $2.20$ \\
\\
$y = \frac{-3.94 + \frac{3.95}{x_{0}^{3}}}{x_{0}^{8}}$ & $12$ & $0.000123$ & $0.0866$ \\
\\
$y = \frac{-2.88 + \frac{2.93}{x_{0}^{4}}}{x_{0}^{7}}$ & $13$ & $8.96 \cdot 10^{-5}$ & $0.319$ \\
\\
$\begin{aligned} y &= \frac{x_{0} - 3.10 + \frac{2.13}{x_{0}^{6}}}{x_{0}^{6}} \end{aligned}$ & $15$ & $3.30 \cdot 10^{-5}$ & $0.499$ \\
\\
$\begin{aligned} y &= \frac{x_{0} - 3.10 + \frac{2.13}{x_{0}^{6}}}{x_{0}^{6}} \end{aligned}$ & $17$ & $3.30 \cdot 10^{-5}$ & $4.84 \cdot 10^{-7}$ \\
\\
$\begin{aligned} y &= 0.00307 + \frac{x_{0} - 3.10 + \frac{2.13}{x_{0}^{6}}}{x_{0}^{6}} \end{aligned}$ & $18$ & $2.59 \cdot 10^{-5}$ & $0.244$ \\
\\
$\begin{aligned} y &= \frac{-11.5 + \frac{11.6}{x_{0}^{2}}}{x_{0}^{3} \left(x_{0}^{6} + x_{0}^{4} - 0.0368\right)} \end{aligned}$ & $19$ & $2.05 \cdot 10^{-5}$ & $0.232$ \\
\\
\bottomrule
\end{tabular}
\end{center}
\end{table}

\begin{table}[h]
\caption{LJ potential for 0-1 interaction || Power 2, 3}
\begin{center}
\begin{tabular}{@{}cccc@{}}
\toprule
Equation & Complexity & Loss & Score \\
\\
\midrule
$y = -0.136$ & $1$ & $0.0607$ & $0.0$ \\
\\
$y = - \frac{0.238}{x_{0}}$ & $3$ & $0.0477$ & $0.121$ \\
\\
$y = - \frac{0.294}{x_{0}^{2}}$ & $4$ & $0.0403$ & $0.169$ \\
\\
\bottomrule
\end{tabular}
\end{center}
\end{table}

\begin{table}[h]
\caption{LJ potential for 0-1 interaction || Power 2, 3, 4}
\begin{center}
\begin{tabular}{@{}cccc@{}}
\toprule
Equation & Complexity & Loss & Score \\
\\
\midrule
$y = -0.136$ & $1$ & $0.0607$ & $0.0$ \\
\\
$y = - \frac{0.238}{x_{0}}$ & $3$ & $0.0477$ & $0.121$ \\
\\
$y = - \frac{0.294}{x_{0}^{2}}$ & $4$ & $0.0403$ & $0.169$ \\
\\
$y = 0.385 - \frac{0.730}{x_{0}}$ & $5$ & $0.0348$ & $0.147$ \\
\\
$y = - \frac{0.778}{x_{0}^{4} + 1.22}$ & $6$ & $0.0342$ & $0.0164$ \\
\\
$y = - \frac{12.1}{x_{0}^{16} + 25.3}$ & $7$ & $0.0267$ & $0.248$ \\
\\
$y = \frac{-2.91 + \frac{2.33}{x_{0}}}{x_{0}^{8}}$ & $9$ & $3.47 \cdot 10^{-5}$ & $3.32$ \\
\\
$\begin{aligned} y &= 0.00259 + \frac{-2.92 + \frac{2.34}{x_{0}}}{x_{0}^{8}} \end{aligned}$ & $11$ & $2.99 \cdot 10^{-5}$ & $0.0736$ \\
\\
$\begin{aligned} y &= 0.00985 + \frac{-0.783 + \frac{0.205}{x_{0}^{6}}}{x_{0}^{6}} \end{aligned}$ & $13$ & $7.81 \cdot 10^{-6}$ & $0.672$ \\
\\
$\begin{aligned} y &= 0.0121 + \frac{-0.783 + \frac{0.205}{x_{0}^{6}}}{x_{0}^{6}} \end{aligned}$ & $14$ & $6.82 \cdot 10^{-6}$ & $0.136$ \\
\\
$\begin{aligned} y &= 0.0112 + \frac{-0.783 + \frac{0.205}{x_{0}^{6}}}{x_{0}^{6}} \end{aligned}$ & $15$ & $5.99 \cdot 10^{-6}$ & $0.129$ \\
\\
$\begin{aligned} y &= 0.00761 x_{0} + \frac{-0.783 + \frac{0.205}{x_{0}^{6}}}{x_{0}^{6}} \end{aligned}$ & $18$ & $4.79 \cdot 10^{-6}$ & $0.0749$ \\
\\
$\begin{aligned} y &= 0.00714 x_{0} + \frac{-0.783 + \frac{0.205}{x_{0}^{6}}}{x_{0}^{6}} \end{aligned}$ & $19$ & $4.46 \cdot 10^{-6}$ & $0.0717$ \\
\\
$\begin{aligned} y &= 0.00725 x_{0} + \frac{-0.783 + \frac{0.205}{x_{0}^{6}}}{x_{0}^{6}} \end{aligned}$ & $20$ & $4.44 \cdot 10^{-6}$ & $0.00449$ \\
\\
\bottomrule
\end{tabular}
\end{center}
\end{table}

\begin{table}[h]
\caption{LJ potential for 0-1 interaction || Power 2, 3, 5}
\begin{center}
\begin{tabular}{@{}cccc@{}}
\toprule
Equation & Complexity & Loss & Score \\
\\
\midrule
$y = -0.136$ & $1$ & $0.0607$ & $0.0$ \\
\\
$y = - \frac{0.238}{x_{0}}$ & $3$ & $0.0477$ & $0.121$ \\
\\
$y = - \frac{0.294}{x_{0}^{2}}$ & $4$ & $0.0403$ & $0.169$ \\
\\
$y = 0.478 x_{0} - 0.870$ & $5$ & $0.0331$ & $0.198$ \\
\\
$y = - \frac{1.08}{x_{0}^{5} + 1.94}$ & $6$ & $0.0320$ & $0.0320$ \\
\\
$y = - \frac{1.68}{x_{0}^{9} + 3.00}$ & $7$ & $0.0294$ & $0.0844$ \\
\\
$y = \frac{-2.63 + \frac{2.08}{x_{0}}}{x_{0}^{5}}$ & $8$ & $0.0132$ & $0.804$ \\
\\
$y = \frac{-2.67 + \frac{2.14}{x_{0}}}{x_{0}^{9}}$ & $9$ & $0.000697$ & $2.94$ \\
\\
$y = \frac{-2.91 + \frac{2.33}{x_{0}}}{x_{0}^{8}}$ & $10$ & $3.47 \cdot 10^{-5}$ & $3.00$ \\
\\
$y = \frac{-2.91 + \frac{2.33}{x_{0}}}{x_{0}^{8}}$ & $12$ & $3.47 \cdot 10^{-5}$ & $1.15 \cdot 10^{-7}$ \\
\\
$\begin{aligned} y &= 0.00258 + \frac{-2.92 + \frac{2.34}{x_{0}}}{x_{0}^{8}} \end{aligned}$ & $13$ & $2.99 \cdot 10^{-5}$ & $0.147$ \\
\\
$\begin{aligned} y &= 0.00258 + \frac{-2.92 + \frac{2.34}{x_{0}}}{x_{0}^{8}} \end{aligned}$ & $14$ & $2.99 \cdot 10^{-5}$ & $3.94 \cdot 10^{-6}$ \\
\\
$\begin{aligned} y &= \frac{0.963 \left(-2.91 + \frac{2.33}{x_{0}}\right)}{x_{0}^{6} \left(x_{0}^{2} - 0.0309\right)} \end{aligned}$ & $16$ & $2.78 \cdot 10^{-5}$ & $0.0370$ \\
\\
$\begin{aligned} y &= \frac{x_{0}^{2} \left(-2.91 + \frac{2.33}{x_{0}}\right)}{\left(x_{0}^{5} + \left(x_{0} - 0.747\right)^{25}\right)^{2}} \end{aligned}$ & $17$ & $2.45 \cdot 10^{-5}$ & $0.125$ \\
\\
$\begin{aligned} y &= \frac{x_{0}^{2} \left(-2.91 + \frac{2.33}{x_{0}}\right)}{\left(x_{0}^{5} + \left(x_{0} - 0.747\right)^{20}\right)^{2}} \end{aligned}$ & $18$ & $2.40 \cdot 10^{-5}$ & $0.0205$ \\
\\
$\begin{aligned} y &= \frac{-2.91 + \frac{2.33}{x_{0}}}{x_{0}^{2} \left(x_{0} + 0.0281\right) \left(x_{0}^{5} - 0.0179\right)} \end{aligned}$ & $19$ & $2.33 \cdot 10^{-5}$ & $0.0301$ \\
\\
\bottomrule
\end{tabular}
\end{center}
\end{table}

\begin{table}[h]
\caption{LJ potential for 0-1 interaction || Power 2, 3, 6}
\begin{center}
\begin{tabular}{@{}cccc@{}}
\toprule
Equation & Complexity & Loss & Score \\
\\
\midrule
$y = -0.136$ & $1$ & $0.0607$ & $0.0$ \\
\\
$y = - \frac{0.238}{x_{0}}$ & $3$ & $0.0477$ & $0.121$ \\
\\
$y = - \frac{0.294}{x_{0}^{2}}$ & $4$ & $0.0403$ & $0.169$ \\
\\
$y = 0.385 - \frac{0.731}{x_{0}}$ & $5$ & $0.0348$ & $0.147$ \\
\\
$y = - \frac{1.23}{x_{0}^{6} + 2.32}$ & $6$ & $0.0307$ & $0.126$ \\
\\
$y = - \frac{2.54}{x_{0}^{9} + 4.93}$ & $7$ & $0.0279$ & $0.0930$ \\
\\
$y = \frac{2.14 - 2.67 x_{0}}{x_{0}^{6}}$ & $8$ & $0.0138$ & $0.707$ \\
\\
$y = \frac{2.33 - 2.91 x_{0}}{x_{0}^{9}}$ & $9$ & $3.47 \cdot 10^{-5}$ & $5.98$ \\
\\
$\begin{aligned} y &= 0.0118 + \frac{-0.786 + \frac{0.205}{x_{0}^{6}}}{x_{0}^{6}} \end{aligned}$ & $13$ & $2.71 \cdot 10^{-6}$ & $0.637$ \\
\\
$\begin{aligned} y &= 0.0118 + \frac{-0.786 + \frac{0.205}{x_{0}^{6}}}{x_{0}^{6}} \end{aligned}$ & $15$ & $2.71 \cdot 10^{-6}$ & $8.49 \cdot 10^{-7}$ \\
\\
$\begin{aligned} y &= 0.0118 + \frac{1.01 \left(-0.786 + \frac{0.205}{x_{0}^{6}}\right)}{x_{0}^{6}} \end{aligned}$ & $17$ & $1.29 \cdot 10^{-6}$ & $0.371$ \\
\\
\bottomrule
\end{tabular}
\end{center}
\end{table}

\begin{table}[h]
\caption{LJ potential for 0-1 interaction || Power 2, 3, 7}
\begin{center}
\begin{tabular}{@{}cccc@{}}
\toprule
Equation & Complexity & Loss & Score \\
\\
\midrule
$y = -0.136$ & $1$ & $0.0607$ & $0.0$ \\
\\
$y = - \frac{0.238}{x_{0}}$ & $3$ & $0.0477$ & $0.121$ \\
\\
$y = - \frac{0.294}{x_{0}^{2}}$ & $4$ & $0.0403$ & $0.169$ \\
\\
$y = 0.385 - \frac{0.730}{x_{0}}$ & $5$ & $0.0348$ & $0.147$ \\
\\
$y = - \frac{1.21}{x_{0}^{7} + 2.32}$ & $6$ & $0.0305$ & $0.133$ \\
\\
$y = - \frac{8.83}{x_{0}^{14} + 17.7}$ & $7$ & $0.0267$ & $0.130$ \\
\\
$\begin{aligned} y &= \frac{-0.764 + \frac{0.154}{x_{0}^{7}}}{x_{0}^{7}} \end{aligned}$ & $9$ & $0.00484$ & $0.854$ \\
\\
$\begin{aligned} y &= \frac{-0.764 + \frac{0.154}{x_{0}^{7}}}{x_{0}^{6}} \end{aligned}$ & $10$ & $0.00149$ & $1.18$ \\
\\
$\begin{aligned} y &= \frac{-1.23 + \frac{0.0543}{x_{0}^{14}}}{x_{0}^{7} + x_{0}} \end{aligned}$ & $12$ & $0.000210$ & $0.982$ \\
\\
$\begin{aligned} y &= \frac{-1.17 + \frac{0.0506}{x_{0}^{14}}}{x_{0}^{7} + x_{0}^{2}} \end{aligned}$ & $13$ & $0.000115$ & $0.599$ \\
\\
$\begin{aligned} y &= \frac{-1.23 + \frac{0.0543}{x_{0}^{14}}}{x_{0}^{8} + x_{0}} \end{aligned}$ & $14$ & $5.01 \cdot 10^{-5}$ & $0.832$ \\
\\
$\begin{aligned} y &= \frac{-1.37 + \frac{0.0601}{x_{0}^{14}}}{x_{0} \left(x_{0}^{7} + x_{0} + 0.285\right)} \end{aligned}$ & $18$ & $3.28 \cdot 10^{-5}$ & $0.106$ \\
\\
\bottomrule
\end{tabular}
\end{center}
\end{table}

\begin{table}[h]
\caption{LJ potential for 0-1 interaction || Power 2, 3, 8}
\begin{center}
\begin{tabular}{@{}cccc@{}}
\toprule
Equation & Complexity & Loss & Score \\
\\
\midrule
$y = -0.136$ & $1$ & $0.0607$ & $0.0$ \\
\\
$y = - \frac{0.238}{x_{0}}$ & $3$ & $0.0477$ & $0.121$ \\
\\
$y = - \frac{0.294}{x_{0}^{2}}$ & $4$ & $0.0403$ & $0.169$ \\
\\
$y = 0.463 x_{0} - 0.833$ & $5$ & $0.0329$ & $0.204$ \\
\\
$y = - \frac{1.23}{x_{0}^{8} + 2.20}$ & $6$ & $0.0306$ & $0.0719$ \\
\\
$y = - \frac{2.67}{x_{0}^{8} + 5.32}$ & $7$ & $0.0282$ & $0.0802$ \\
\\
$y = \frac{-2.91 + \frac{2.33}{x_{0}}}{x_{0}^{8}}$ & $8$ & $3.47 \cdot 10^{-5}$ & $6.70$ \\
\\
$\begin{aligned} y &= 0.00118 + \frac{-2.91 + \frac{2.33}{x_{0}}}{x_{0}^{8}} \end{aligned}$ & $10$ & $3.17 \cdot 10^{-5}$ & $0.0444$ \\
\\
$\begin{aligned} y &= \frac{-2.91 + \frac{2.33}{x_{0}}}{x_{0}^{8} - 0.00250} \end{aligned}$ & $11$ & $3.00 \cdot 10^{-5}$ & $0.0564$ \\
\\
$\begin{aligned} y &= \frac{-2.91 + \frac{2.33}{x_{0}}}{x_{0}^{8} + \left(x_{0} - 0.956\right)^{3}} \end{aligned}$ & $13$ & $2.66 \cdot 10^{-5}$ & $0.0598$ \\
\\
$\begin{aligned} y &= \frac{-2.91 + \frac{2.33}{x_{0}}}{x_{0}^{8} + \left(x_{0} - 0.605\right)^{24}} \end{aligned}$ & $14$ & $2.40 \cdot 10^{-5}$ & $0.103$ \\
\\
$\begin{aligned} y &= \frac{1.04 \left(-2.91 + \frac{2.33}{x_{0}}\right)}{x_{0}^{8} + \left(x_{0} - 0.647\right)^{3}} \end{aligned}$ & $15$ & $2.35 \cdot 10^{-5}$ & $0.0219$ \\
\\
$\begin{aligned} y &= \frac{-2.91 + \frac{2.33}{x_{0}}}{x_{0}^{8} + \left(\left(x_{0} - 0.647\right)^{9} - 0.170\right)^{3}} \end{aligned}$ & $17$ & $2.33 \cdot 10^{-5}$ & $0.00480$ \\
\\
$\begin{aligned} y &= \frac{-2.91 + \frac{2.33}{x_{0}}}{x_{0}^{8} + \left(\left(x_{0} - 0.647\right)^{8} - 0.103\right)^{3}} \end{aligned}$ & $18$ & $2.13 \cdot 10^{-5}$ & $0.0878$ \\
\\
\bottomrule
\end{tabular}
\end{center}
\end{table}

\begin{table}[h]
\caption{LJ potential for 0-1 interaction || Power 2, 3, 9}
\begin{center}
\begin{tabular}{@{}cccc@{}}
\toprule
Equation & Complexity & Loss & Score \\
\\
\midrule
$y = -0.136$ & $1$ & $0.0607$ & $0.0$ \\
\\
$y = - \frac{0.238}{x_{0}}$ & $3$ & $0.0477$ & $0.121$ \\
\\
$y = - \frac{0.294}{x_{0}^{2}}$ & $4$ & $0.0403$ & $0.169$ \\
\\
$y = 0.453 x_{0} - 0.821$ & $5$ & $0.0328$ & $0.205$ \\
\\
$y = - \frac{1.65}{x_{0}^{9} + 2.91}$ & $6$ & $0.0295$ & $0.106$ \\
\\
$y = - \frac{2.91}{x_{0}^{9} + 5.74}$ & $7$ & $0.0277$ & $0.0634$ \\
\\
$y = \frac{-2.67 + \frac{2.14}{x_{0}}}{x_{0}^{9}}$ & $8$ & $0.000697$ & $3.68$ \\
\\
$y = \frac{-2.91 + \frac{2.33}{x_{0}}}{x_{0}^{8}}$ & $10$ & $3.47 \cdot 10^{-5}$ & $1.50$ \\
\\
$y = \frac{-2.91 + \frac{2.33}{x_{0}}}{x_{0}^{8}}$ & $11$ & $3.47 \cdot 10^{-5}$ & $6.34 \cdot 10^{-7}$ \\
\\
$\begin{aligned} y &= \frac{-2.91 + \frac{2.33}{x_{0}}}{x_{0}^{8} - 0.00253} \end{aligned}$ & $13$ & $3.00 \cdot 10^{-5}$ & $0.0726$ \\
\\
$\begin{aligned} y &= \frac{-2.91 + \frac{2.33}{x_{0}}}{x_{0}^{8} - \frac{0.00253}{x_{0}^{3}}} \end{aligned}$ & $15$ & $2.91 \cdot 10^{-5}$ & $0.0143$ \\
\\
$\begin{aligned} y &= \frac{-2.91 + \frac{2.33}{x_{0}}}{0.000497 x_{0}^{18} + x_{0}^{8}} \end{aligned}$ & $16$ & $2.91 \cdot 10^{-5}$ & $0.00117$ \\
\\
$\begin{aligned} y &= \frac{-2.91 + \frac{2.33}{x_{0}}}{x_{0}^{6} \left(x_{0} + \left(x_{0} - 0.834\right)^{9}\right)^{2}} \end{aligned}$ & $17$ & $2.85 \cdot 10^{-5}$ & $0.0207$ \\
\\
$\begin{aligned} y &= \frac{-2.91 + \frac{2.33}{x_{0}}}{x_{0}^{7} \left(x_{0} + \left(x_{0} - 0.834\right)^{9}\right)} \end{aligned}$ & $18$ & $2.49 \cdot 10^{-5}$ & $0.137$ \\
\\
$\begin{aligned} y &= \frac{-2.91 + \frac{2.33}{x_{0}}}{x_{0}^{7} \left(x_{0} + \left(x_{0} - 0.834\right)^{18}\right)} \end{aligned}$ & $19$ & $2.42 \cdot 10^{-5}$ & $0.0284$ \\
\\
\bottomrule
\end{tabular}
\end{center}
\end{table}

\begin{table}[h]
\caption{LJ potential for 0-1 interaction || Power 2, 3, 10}
\begin{center}
\begin{tabular}{@{}cccc@{}}
\toprule
Equation & Complexity & Loss & Score \\
\\
\midrule
$y = -0.136$ & $1$ & $0.0607$ & $0.0$ \\
\\
$y = - \frac{0.238}{x_{0}}$ & $3$ & $0.0477$ & $0.121$ \\
\\
$y = - \frac{0.294}{x_{0}^{2}}$ & $4$ & $0.0403$ & $0.169$ \\
\\
$y = 0.385 - \frac{0.731}{x_{0}}$ & $5$ & $0.0348$ & $0.147$ \\
\\
$y = - \frac{3.27}{x_{0}^{10} + 5.73}$ & $6$ & $0.0282$ & $0.208$ \\
\\
$y = - \frac{3.27}{x_{0}^{10} + \frac{5.73}{x_{0}}}$ & $8$ & $0.0242$ & $0.0771$ \\
\\
$\begin{aligned} y &= \frac{-1.41 + \frac{0.902}{x_{0}^{2}}}{x_{0}^{10}} \end{aligned}$ & $9$ & $0.00504$ & $1.57$ \\
\\
$\begin{aligned} y &= \frac{-0.567 + \frac{0.0597}{x_{0}^{10}}}{x_{0}^{6}} \end{aligned}$ & $10$ & $0.00100$ & $1.62$ \\
\\
$\begin{aligned} y &= \frac{-0.773 + \frac{0.203}{x_{0}^{6}}}{x_{0}^{6}} \end{aligned}$ & $11$ & $0.000109$ & $2.22$ \\
\\
$\begin{aligned} y &= \frac{0.0489 + \frac{- \frac{0.702}{x_{0}} + \frac{0.0915}{x_{0}^{10}}}{x_{0}^{3}}}{x_{0}} \end{aligned}$ & $15$ & $2.88 \cdot 10^{-5}$ & $0.333$ \\
\\
$\begin{aligned} y &= \frac{-0.811 + \frac{0.0879}{x_{0}^{10}}}{x_{0}^{3} \cdot \left(0.267 x_{0}^{6} + x_{0}\right)} \end{aligned}$ & $18$ & $2.80 \cdot 10^{-5}$ & $0.00891$ \\
\\
$\begin{aligned} y &= \frac{-0.818 + \frac{0.0879}{x_{0}^{10}}}{x_{0}^{3} \cdot \left(0.279 x_{0}^{6} + x_{0}\right)} \end{aligned}$ & $20$ & $1.33 \cdot 10^{-5}$ & $0.374$ \\
\\
\bottomrule
\end{tabular}
\end{center}
\end{table}

\begin{table}[h]
\caption{LJ potential for 1-1 interaction || Power 2, 3}
\begin{center}
\begin{tabular}{@{}cccc@{}}
\toprule
Equation & Complexity & Loss & Score \\
\\
\midrule
$y = -0.0132$ & $1$ & $0.0323$ & $0.0$ \\
\\
$y = \frac{0.0917}{x_{0}^{3}}$ & $4$ & $0.0307$ & $0.0171$ \\
\\
$y = \frac{0.124}{x_{0}^{9}}$ & $5$ & $0.0152$ & $0.701$ \\
\\
$y = \frac{0.00327}{x_{0}^{27}}$ & $6$ & $0.00508$ & $1.10$ \\
\\
$y = \frac{0.000443}{x_{0}^{36}}$ & $7$ & $0.00492$ & $0.0302$ \\
\\
$y = -0.0322 + \frac{0.00330}{x_{0}^{27}}$ & $8$ & $0.00378$ & $0.265$ \\
\\
$y = -0.0349 + \frac{0.000449}{x_{0}^{36}}$ & $9$ & $0.00371$ & $0.0188$ \\
\\
$\begin{aligned} y &= \frac{-0.264 + \frac{0.117}{x_{0}^{9}}}{x_{0}^{3}} \end{aligned}$ & $10$ & $0.00258$ & $0.361$ \\
\\
$\begin{aligned} y &= \frac{-0.464 + \frac{0.215}{x_{0}^{6}}}{x_{0}^{6}} \end{aligned}$ & $11$ & $1.16 \cdot 10^{-5}$ & $5.41$ \\
\\
$\begin{aligned} y &= \frac{-0.464 + \frac{0.215}{x_{0}^{6}}}{\left(x_{0}^{3} - 0.000990\right)^{2}} \end{aligned}$ & $15$ & $1.15 \cdot 10^{-5}$ & $0.00231$ \\
\\
$\begin{aligned} y &= \frac{- \frac{0.625}{x_{0}^{2}} + \frac{0.373}{x_{0}^{6}}}{x_{0}^{5}} \end{aligned}$ & $16$ & $8.13 \cdot 10^{-6}$ & $0.344$ \\
\\
$\begin{aligned} y &= \frac{-0.464 + \frac{0.215}{x_{0}^{6}}}{\left(0.00578 x_{0}^{9} + x_{0}^{3}\right)^{2}} \end{aligned}$ & $17$ & $4.52 \cdot 10^{-6}$ & $0.587$ \\
\\
$\begin{aligned} y &= \frac{-0.464 + \frac{0.215}{x_{0}^{6}}}{\left(0.00104 x_{0}^{12} + x_{0}^{3}\right)^{2}} \end{aligned}$ & $18$ & $1.13 \cdot 10^{-6}$ & $1.39$ \\
\\
$\begin{aligned} y &= \frac{-0.464 + \frac{0.215}{x_{0}^{6}}}{\left(0.00104 x_{0}^{12} + x_{0}^{3}\right)^{2}} \end{aligned}$ & $19$ & $1.13 \cdot 10^{-6}$ & $1.77 \cdot 10^{-7}$ \\
\\
$\begin{aligned} y &= \frac{-0.464 + \frac{0.215}{x_{0}^{6}}}{\left(0.000886 x_{0}^{12} + x_{0}^{3}\right)^{2}} \end{aligned}$ & $20$ & $1.07 \cdot 10^{-6}$ & $0.0533$ \\
\\
\bottomrule
\end{tabular}
\end{center}
\end{table}

\begin{table}[h]
\caption{LJ potential for 1-1 interaction || Power 2, 3, 4}
\begin{center}
\begin{tabular}{@{}cccc@{}}
\toprule
Equation & Complexity & Loss & Score \\
\\
\midrule
$y = -0.0132$ & $1$ & $0.0323$ & $0.0$ \\
\\
$y = -0.0132$ & $3$ & $0.0323$ & $3.09 \cdot 10^{-8}$ \\
\\
$y = 6.41 \left(0.628 x_{0} - 1\right)^{4}$ & $4$ & $0.0256$ & $0.232$ \\
\\
\bottomrule
\end{tabular}
\end{center}
\end{table}

\begin{table}[h]
\caption{LJ potential for 1-1 interaction || Power 2, 3, 5}
\begin{center}
\begin{tabular}{@{}cccc@{}}
\toprule
Equation & Complexity & Loss & Score \\
\\
\midrule
$y = -0.0132$ & $1$ & $0.0323$ & $0.0$ \\
\\
$y = \frac{0.158}{x_{0}^{5}}$ & $4$ & $0.0253$ & $0.0818$ \\
\\
$y = \frac{0.00508}{x_{0}^{25}}$ & $5$ & $0.00521$ & $1.58$ \\
\\
$y = \frac{0.00168}{x_{0}^{30}}$ & $6$ & $0.00497$ & $0.0472$ \\
\\
$y = -0.0346 + \frac{0.00534}{x_{0}^{25}}$ & $7$ & $0.00386$ & $0.251$ \\
\\
$y = -0.0355 + \frac{0.00176}{x_{0}^{30}}$ & $8$ & $0.00370$ & $0.0441$ \\
\\
$y = \frac{-1.97 + \frac{1.73}{x_{0}}}{x_{0}^{9}}$ & $9$ & $0.000136$ & $3.30$ \\
\\
$y = \frac{-2.20 + \frac{1.94}{x_{0}}}{x_{0}^{8}}$ & $10$ & $1.60 \cdot 10^{-5}$ & $2.14$ \\
\\
$y = \frac{-2.20 + \frac{1.94}{x_{0}}}{x_{0}^{8}}$ & $11$ & $1.60 \cdot 10^{-5}$ & $4.38 \cdot 10^{-7}$ \\
\\
$y = \frac{-2.20 + \frac{1.94}{x_{0}}}{x_{0}^{8}}$ & $12$ & $1.60 \cdot 10^{-5}$ & $4.31 \cdot 10^{-6}$ \\
\\
$y = \frac{-2.20 + \frac{1.94}{x_{0}}}{x_{0}^{8}}$ & $13$ & $1.60 \cdot 10^{-5}$ & $3.13 \cdot 10^{-6}$ \\
\\
$\begin{aligned} y &= \frac{-2.19 + \frac{1.93}{x_{0}}}{x_{0}^{2} \left(x_{0}^{6} - 0.00670\right)} \end{aligned}$ & $14$ & $1.23 \cdot 10^{-5}$ & $0.263$ \\
\\
$\begin{aligned} y &= \frac{-2.19 + \frac{1.93}{x_{0}}}{x_{0}^{2} \left(x_{0}^{6} - 0.00584\right)} \end{aligned}$ & $15$ & $1.19 \cdot 10^{-5}$ & $0.0342$ \\
\\
$\begin{aligned} y &= \frac{-2.19 + \frac{1.93}{x_{0}}}{x_{0}^{2} \left(x_{0}^{6} - 0.00584\right)} \end{aligned}$ & $17$ & $1.19 \cdot 10^{-5}$ & $2.10 \cdot 10^{-7}$ \\
\\
$\begin{aligned} y &= \frac{-2.19 + \frac{1.93}{x_{0}}}{x_{0}^{2} \left(x_{0}^{6} - \frac{0.000213}{x_{0}^{15}}\right)} \end{aligned}$ & $18$ & $9.78 \cdot 10^{-6}$ & $0.195$ \\
\\
$\begin{aligned} y &= \frac{-2.19 + \frac{1.93}{x_{0}}}{x_{0}^{2} \left(x_{0}^{6} - \frac{0.000213}{x_{0}^{15}}\right)} \end{aligned}$ & $19$ & $9.78 \cdot 10^{-6}$ & $5.11 \cdot 10^{-7}$ \\
\\
\bottomrule
\end{tabular}
\end{center}
\end{table}

\begin{table}[h]
\caption{LJ potential for 1-1 interaction || Power 2, 3, 6}
\begin{center}
\begin{tabular}{@{}cccc@{}}
\toprule
Equation & Complexity & Loss & Score \\
\\
\midrule
$y = -0.0132$ & $1$ & $0.0323$ & $0.0$ \\
\\
$y = \frac{0.162}{x_{0}^{6}}$ & $4$ & $0.0224$ & $0.122$ \\
\\
$y = \frac{0.000443}{x_{0}^{36}}$ & $5$ & $0.00492$ & $1.51$ \\
\\
$y = -0.0353 + \frac{0.000453}{x_{0}^{36}}$ & $7$ & $0.00370$ & $0.143$ \\
\\
$\begin{aligned} y &= \frac{-0.464 + \frac{0.215}{x_{0}^{6}}}{x_{0}^{6}} \end{aligned}$ & $9$ & $1.16 \cdot 10^{-5}$ & $2.88$ \\
\\
$\begin{aligned} y &= \frac{-0.464 + \frac{0.215}{x_{0}^{6}}}{x_{0}^{6}} \end{aligned}$ & $10$ & $1.16 \cdot 10^{-5}$ & $2.59 \cdot 10^{-7}$ \\
\\
$\begin{aligned} y &= 0.00379 + \frac{-0.471 + \frac{0.217}{x_{0}^{6}}}{x_{0}^{6}} \end{aligned}$ & $11$ & $5.75 \cdot 10^{-7}$ & $3.00$ \\
\\
$\begin{aligned} y &= 0.00379 + \frac{-0.471 + \frac{1.43 \cdot 10^{-5}}{x_{0}^{2}} + \frac{0.217}{x_{0}^{6}}}{x_{0}^{6}} \end{aligned}$ & $16$ & $5.75 \cdot 10^{-7}$ & $5.28 \cdot 10^{-5}$ \\
\\
$\begin{aligned} y &= 0.00379 + \frac{-0.471 + \frac{0.217}{x_{0}^{6}} - \frac{3.42 \cdot 10^{-6}}{x_{0}^{9}}}{x_{0}^{6}} \end{aligned}$ & $17$ & $5.75 \cdot 10^{-7}$ & $0.000248$ \\
\\
$\begin{aligned} y &= 0.00379 + \frac{-0.471 + \frac{0.217}{x_{0}^{6}} - \frac{3.42 \cdot 10^{-6}}{x_{0}^{18}}}{x_{0}^{6}} \end{aligned}$ & $18$ & $5.72 \cdot 10^{-7}$ & $0.00494$ \\
\\
$\begin{aligned} y &= 0.00379 + \frac{-0.471 + \frac{0.217}{x_{0}^{6}} - \frac{2.76 \cdot 10^{-8}}{x_{0}^{36}}}{x_{0}^{6}} \end{aligned}$ & $19$ & $5.69 \cdot 10^{-7}$ & $0.00509$ \\
\\
$\begin{aligned} y &= 0.00379 + \frac{-0.471 + \frac{0.217}{x_{0}^{6}} - \frac{3.78 \cdot 10^{-7}}{x_{0}^{27}}}{x_{0}^{6}} \end{aligned}$ & $20$ & $5.68 \cdot 10^{-7}$ & $0.00180$ \\
\\
\bottomrule
\end{tabular}
\end{center}
\end{table}

\begin{table}[h]
\caption{LJ potential for 1-1 interaction || Power 2, 3, 7}
\begin{center}
\begin{tabular}{@{}cccc@{}}
\toprule
Equation & Complexity & Loss & Score \\
\\
\midrule
$y = -0.0132$ & $1$ & $0.0323$ & $0.0$ \\
\\
$y = \frac{0.155}{x_{0}^{7}}$ & $4$ & $0.0197$ & $0.165$ \\
\\
$y = \frac{2.43 \cdot 10^{-5}}{x_{0}^{49}}$ & $5$ & $0.00501$ & $1.37$ \\
\\
$y = \frac{0.000116}{x_{0}^{42}}$ & $6$ & $0.00496$ & $0.0102$ \\
\\
$y = -0.0386 + \frac{0.0126}{x_{0}^{21}}$ & $7$ & $0.00429$ & $0.145$ \\
\\
$y = -0.0365 + \frac{0.00270}{x_{0}^{28}}$ & $8$ & $0.00372$ & $0.142$ \\
\\
$\begin{aligned} y &= \frac{-0.0690 + \frac{2.07 \cdot 10^{-5}}{x_{0}^{49}}}{x_{0}} \end{aligned}$ & $9$ & $0.00311$ & $0.178$ \\
\\
$\begin{aligned} y &= \frac{-0.231 + \frac{0.0402}{x_{0}^{14}}}{x_{0}^{3}} \end{aligned}$ & $10$ & $0.000924$ & $1.21$ \\
\\
$\begin{aligned} y &= \frac{-0.196 + \frac{0.0161}{x_{0}^{18}}}{x_{0}^{3}} \end{aligned}$ & $11$ & $0.000792$ & $0.154$ \\
\\
$\begin{aligned} y &= \frac{- \frac{0.246}{x_{0}} + \frac{0.0446}{x_{0}^{14}}}{x_{0}^{3}} \end{aligned}$ & $12$ & $0.000291$ & $1.00$ \\
\\
$\begin{aligned} y &= \frac{-0.642 + \frac{0.139}{x_{0}^{12}}}{x_{0}^{7} + x_{0}} \end{aligned}$ & $13$ & $1.94 \cdot 10^{-6}$ & $5.01$ \\
\\
$\begin{aligned} y &= \frac{-0.642 + \frac{0.139}{x_{0}^{12}}}{x_{0}^{7} + x_{0} - 0.00116} \end{aligned}$ & $15$ & $1.89 \cdot 10^{-6}$ & $0.0134$ \\
\\
$\begin{aligned} y &= \frac{-0.642 + \frac{0.139}{x_{0}^{12}}}{x_{0}^{7} + x_{0} - \frac{0.00116}{x_{0}}} \end{aligned}$ & $17$ & $1.88 \cdot 10^{-6}$ & $0.00116$ \\
\\
$\begin{aligned} y &= \frac{-0.642 + \frac{0.139}{x_{0}^{12}}}{x_{0}^{7} + x_{0} + \left(x_{0} - 0.707\right)^{14}} \end{aligned}$ & $19$ & $1.19 \cdot 10^{-6}$ & $0.229$ \\
\\
$\begin{aligned} y &= \frac{-0.642 + \frac{0.139}{x_{0}^{12}}}{x_{0}^{7} + x_{0} + \left(x_{0} - 0.901\right)^{28}} \end{aligned}$ & $20$ & $1.10 \cdot 10^{-6}$ & $0.0835$ \\
\\
\bottomrule
\end{tabular}
\end{center}
\end{table}

\begin{table}[h]
\caption{LJ potential for 1-1 interaction || Power 2, 3, 8}
\begin{center}
\begin{tabular}{@{}cccc@{}}
\toprule
Equation & Complexity & Loss & Score \\
\\
\midrule
$y = -0.0132$ & $1$ & $0.0323$ & $0.0$ \\
\\
$y = - 0.00456 x_{0}$ & $3$ & $0.0323$ & $0.000124$ \\
\\
$y = 95.2 \left(0.566 x_{0} - 1\right)^{8}$ & $4$ & $0.0170$ & $0.640$ \\
\\
$y = \frac{8.54 \cdot 10^{-7}}{x_{0}^{64}}$ & $5$ & $0.00508$ & $1.21$ \\
\\
$y = \frac{0.00108}{x_{0}^{32}}$ & $6$ & $0.00494$ & $0.0281$ \\
\\
$y = -0.0375 + \frac{0.00644}{x_{0}^{24}}$ & $7$ & $0.00392$ & $0.230$ \\
\\
$y = -0.0347 + \frac{0.00108}{x_{0}^{32}}$ & $8$ & $0.00370$ & $0.0588$ \\
\\
$\begin{aligned} y &= \frac{-0.0709 + \frac{0.00530}{x_{0}^{24}}}{x_{0}} \end{aligned}$ & $9$ & $0.00285$ & $0.260$ \\
\\
$\begin{aligned} y &= \frac{-0.154 + \frac{0.0231}{x_{0}^{16}}}{x_{0}^{3}} \end{aligned}$ & $10$ & $0.00114$ & $0.919$ \\
\\
$\begin{aligned} y &= \frac{-0.463 + \frac{0.215}{x_{0}^{6}}}{x_{0}^{6}} \end{aligned}$ & $11$ & $1.16 \cdot 10^{-5}$ & $4.59$ \\
\\
$\begin{aligned} y &= 0.00657 + \frac{-0.474 + \frac{0.217}{x_{0}^{6}}}{x_{0}^{6}} \end{aligned}$ & $13$ & $6.93 \cdot 10^{-6}$ & $0.257$ \\
\\
$\begin{aligned} y &= 0.00255 + \frac{-0.474 + \frac{0.217}{x_{0}^{6}}}{x_{0}^{6}} \end{aligned}$ & $14$ & $4.29 \cdot 10^{-6}$ & $0.481$ \\
\\
$\begin{aligned} y &= \frac{0.00657}{x_{0}} + \frac{-0.474 + \frac{0.217}{x_{0}^{6}}}{x_{0}^{6}} \end{aligned}$ & $15$ & $1.07 \cdot 10^{-6}$ & $1.39$ \\
\\
$\begin{aligned} y &= \frac{0.00657 + \frac{-0.474 + \frac{0.217}{x_{0}^{6}}}{x_{0}^{5}}}{x_{0}} \end{aligned}$ & $17$ & $1.07 \cdot 10^{-6}$ & $1.40 \cdot 10^{-7}$ \\
\\
$\begin{aligned} y &= \frac{0.00657 + \frac{-0.474 + \frac{0.217}{x_{0}^{6}}}{x_{0}^{5}}}{x_{0}} \end{aligned}$ & $18$ & $1.07 \cdot 10^{-6}$ & $3.36 \cdot 10^{-6}$ \\
\\
\bottomrule
\end{tabular}
\end{center}
\end{table}

\begin{table}[h]
\caption{LJ potential for 1-1 interaction || Power 2, 3, 9}
\begin{center}
\begin{tabular}{@{}cccc@{}}
\toprule
Equation & Complexity & Loss & Score \\
\\
\midrule
$y = -0.0132$ & $1$ & $0.0323$ & $0.0$ \\
\\
$y = \frac{0.124}{x_{0}^{9}}$ & $4$ & $0.0152$ & $0.251$ \\
\\
$y = \frac{0.00327}{x_{0}^{27}}$ & $5$ & $0.00508$ & $1.10$ \\
\\
$y = \frac{0.000443}{x_{0}^{36}}$ & $6$ & $0.00492$ & $0.0302$ \\
\\
$y = -0.0367 + \frac{0.00336}{x_{0}^{27}}$ & $7$ & $0.00375$ & $0.273$ \\
\\
$y = -0.0355 + \frac{0.000438}{x_{0}^{36}}$ & $8$ & $0.00374$ & $0.00190$ \\
\\
$\begin{aligned} y &= \frac{-0.0758 + \frac{0.00282}{x_{0}^{27}}}{x_{0}} \end{aligned}$ & $9$ & $0.00275$ & $0.307$ \\
\\
$\begin{aligned} y &= \frac{-0.333 + \frac{0.118}{x_{0}^{9}}}{x_{0}^{4}} \end{aligned}$ & $10$ & $0.000640$ & $1.46$ \\
\\
$\begin{aligned} y &= \frac{-0.787 + \frac{0.254}{x_{0}^{9}}}{x_{0}^{9} + x_{0}} \end{aligned}$ & $11$ & $0.000581$ & $0.0978$ \\
\\
$\begin{aligned} y &= \frac{- \frac{0.376}{x_{0}} + \frac{0.138}{x_{0}^{9}}}{x_{0}^{4}} \end{aligned}$ & $12$ & $0.000108$ & $1.68$ \\
\\
$\begin{aligned} y &= \frac{0.225 - \frac{0.620}{x_{0}} + \frac{0.149}{x_{0}^{9}}}{x_{0}^{4}} \end{aligned}$ & $14$ & $8.12 \cdot 10^{-6}$ & $1.30$ \\
\\
$\begin{aligned} y &= \frac{-0.732 + \frac{0.230}{x_{0}^{9}}}{x_{0} \left(x_{0}^{6} + x_{0}^{2}\right)} \end{aligned}$ & $15$ & $2.35 \cdot 10^{-6}$ & $1.24$ \\
\\
\bottomrule
\end{tabular}
\end{center}
\end{table}

\begin{table}[h]
\caption{LJ potential for 1-1 interaction || Power 2, 3, 10}
\begin{center}
\begin{tabular}{@{}cccc@{}}
\toprule
Equation & Complexity & Loss & Score \\
\\
\midrule
$y = -0.0132$ & $1$ & $0.0323$ & $0.0$ \\
\\
$y = -0.0132$ & $2$ & $0.0323$ & $6.19 \cdot 10^{-8}$ \\
\\
$y = \frac{0.107}{x_{0}^{10}}$ & $4$ & $0.0135$ & $0.438$ \\
\\
$y = \frac{0.00168}{x_{0}^{30}}$ & $5$ & $0.00497$ & $0.997$ \\
\\
$y = \frac{0.000181}{x_{0}^{40}}$ & $6$ & $0.00494$ & $0.00512$ \\
\\
$y = -0.0360 + \frac{0.00173}{x_{0}^{30}}$ & $7$ & $0.00369$ & $0.293$ \\
\\
$\begin{aligned} y &= \frac{-0.264 + \frac{0.0992}{x_{0}^{10}}}{x_{0}^{3}} \end{aligned}$ & $9$ & $0.00192$ & $0.325$ \\
\\
$\begin{aligned} y &= \frac{-0.289 + \frac{0.0700}{x_{0}^{10}}}{x_{0}^{6}} \end{aligned}$ & $10$ & $0.000322$ & $1.79$ \\
\\
$\begin{aligned} y &= \frac{- \frac{0.339}{x_{0}} + \frac{0.105}{x_{0}^{10}}}{x_{0}^{4}} \end{aligned}$ & $12$ & $7.07 \cdot 10^{-5}$ & $0.757$ \\
\\
$\begin{aligned} y &= \frac{- \frac{0.339}{x_{0}} + \frac{0.105}{x_{0}^{10}}}{x_{0}^{4}} \end{aligned}$ & $13$ & $7.07 \cdot 10^{-5}$ & $4.24 \cdot 10^{-7}$ \\
\\
$\begin{aligned} y &= \frac{- \frac{0.506}{x_{0}^{3} + 0.322} + \frac{0.140}{x_{0}^{10}}}{x_{0}^{3}} \end{aligned}$ & $14$ & $1.85 \cdot 10^{-5}$ & $1.34$ \\
\\
$\begin{aligned} y &= \frac{- \frac{0.506}{x_{0}^{3} + 0.319} + \frac{0.140}{x_{0}^{10}}}{x_{0}^{3}} \end{aligned}$ & $16$ & $1.84 \cdot 10^{-5}$ & $0.00282$ \\
\\
$\begin{aligned} y &= 0.00385 + \frac{- \frac{0.506}{x_{0}^{3} + 0.322} + \frac{0.140}{x_{0}^{10}}}{x_{0}^{3}} \end{aligned}$ & $17$ & $1.03 \cdot 10^{-5}$ & $0.584$ \\
\\
$\begin{aligned} y &= 0.00274 + \frac{- \frac{0.506}{x_{0}^{3} + 0.322} + \frac{0.140}{x_{0}^{10}}}{x_{0}^{3}} \end{aligned}$ & $18$ & $9.58 \cdot 10^{-6}$ & $0.0678$ \\
\\
\bottomrule
\end{tabular}
\end{center}
\end{table}

\end{document}

%% file: math_commands.tex
\usepackage{amsmath,amsfonts,bm}

\def\eqref#1{equation~\ref{#1}}

\def\1{\bm{1}}

\DeclareMathAlphabet{\mathsfit}{\encodingdefault}{\sfdefault}{m}{sl}
\SetMathAlphabet{\mathsfit}{bold}{\encodingdefault}{\sfdefault}{bx}{n}

\newcommand{\gnn}{\textsc{Gnn}{}{}}
\newcommand{\hnn}{\textsc{Hnn}{}{}}
\newcommand{\hgnn}{\textsc{Hgnn}{}{}}
\newcommand{\hgn}{\textsc{Hgn}{}{}}
\newcommand{\mlp}{\texttt{MLP}{}{}}
\newcommand{\x}{\mathbf{x}}
\newcommand{\p}{\mathbf{p_x}}

